\pdfoutput=1

\documentclass[11pt]{article}

\usepackage[final]{acl}

\usepackage{times}
\usepackage{latexsym}

\usepackage[T1]{fontenc}

\usepackage[utf8]{inputenc}

\usepackage{microtype}

\usepackage{inconsolata}

\usepackage{amsmath,amsfonts,bm}

\def\Figref#1{Figure~\ref{#1}}

\def\Ssecref#1{\S~\ref{#1}}
\def\eqref#1{equation~\ref{#1}}

\def\Tabref#1{Table~\ref{#1}}

\def\1{\bm{1}}

\DeclareMathAlphabet{\mathsfit}{\encodingdefault}{\sfdefault}{m}{sl}
\SetMathAlphabet{\mathsfit}{bold}{\encodingdefault}{\sfdefault}{bx}{n}

\usepackage{booktabs}
\usepackage{url}
\usepackage{times}
\usepackage{latexsym}

\usepackage{amsmath,amssymb}
\usepackage{graphicx}
\usepackage{adjustbox}
\usepackage{tabularx}
\usepackage{ragged2e}
\usepackage{multirow}
\usepackage{pifont}
\usepackage{soul}

\usepackage[toc,page]{appendix}
\usepackage{xcolor}
\usepackage{color}
\usepackage{floatrow}
\usepackage{makecell}
\usepackage{xspace}
\usepackage{colortbl}

\definecolor{airforceblue}{rgb}{0.36, 0.54, 0.66}
\definecolor{ao(english)}{rgb}{0.0, 0.5, 0.0}
\definecolor{auburn}{rgb}{0.43, 0.21, 0.1}
\definecolor{bittersweet}{rgb}{1.0, 0.44, 0.37}

\usepackage[tableposition=top]{caption}
\usepackage{floatrow}
\usepackage{subcaption}
\usepackage{wrapfig}
\usepackage{makecell}

\newfloatcommand{capbtabbox}{table}[][\FBwidth]

\usepackage{arydshln}

\usepackage{minibox}
\usepackage{tabularx,ragged2e}
\usepackage{tabularx} 
\usepackage{algorithm}
\usepackage[noend]{algpseudocode}
\usepackage{bbm}

\newcommand\ti[1]{\textit{#1}}

\newcommand\tf[1]{\textbf{#1}}
\newcommand\ttt[1]{\texttt{#1}}

\newcommand{\lt}{long-tail~}
\newcommand{\Lt}{Long-tail~}
\newcommand{\framework}{LINK}
\newcommand{\dataset}{LINT}
\newcommand{\chat}{\texttt{ChatGPT}}
\newcommand{\gpt}{\texttt{GPT4}}
\newcommand{\instruct}{\texttt{InstructGPT}}

\newcommand{\benchmarkEmoji}{\includegraphics[height=.9em,trim=0 .4em 0 0]{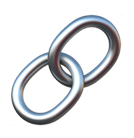}}

\title{In Search of the Long-Tail: Systematic Generation of Long-Tail Inferential Knowledge via Logical Rule Guided Search}

\author{Huihan Li$^{1}$ \hspace{3mm} Yuting Ning$^{2}$ \hspace{3mm} Zeyi Liao$^{2}$ \hspace{3mm} \textbf{Siyuan Wang}$^{3}$ \hspace{3mm} \textbf{Xiang Lorraine Li}$^{4,5}$ \hspace{3mm} \\ \textbf{Ximing Lu}$^{4,6}$ \hspace{3mm} \textbf{Wenting Zhao}$^{7}$ 
\hspace{3mm} \textbf{Faeze Brahman}$^{4}$ \hspace{3mm} \textbf{Yejin Choi}$^{4,6}$ \hspace{3mm} \textbf{Xiang Ren}$^{1,4}$ \vspace{1mm} \\
$^{1}$University of Southern California \hspace{1mm}
$^{2}$ Ohio State University \hspace{1mm} $^{3}$Fudan University \\
$^{4}$Allen Institute for Artificial Intelligence \hspace{1mm} $^{5}$ University of Pittsburgh \\
$^{6}$ University of Washington \hspace{1mm} $^{7}$ Cornell University \\
\small{\texttt{\{huihanl,xiangren\}@usc.edu}}}

\begin{document}
\maketitle
\begin{abstract}
To effectively use large language models (LLMs) for real-world queries, it is imperative that they generalize to the \textit{\lt distribution}, \textit{i.e.}, rare examples where models exhibit low confidence. In this work, we take the first step towards evaluating LLMs in the \lt distribution of inferential knowledge. We exemplify \lt evaluation on the Natural Language Inference task. First, we introduce \tf{L}ogic-\tf{In}duced-\tf{K}nowledge-Search (\tf{\framework}\benchmarkEmoji), a systematic \lt data generation framework, to obtain factually-correct yet \lt inferential statements. \framework\ uses variable-wise prompting grounded on symbolic rules to seek low-confidence statements while ensuring factual correctness. We then use \framework\ to curate \tf{L}ogic-\tf{In}duced-Long-\tf{T}ail (\tf{\dataset}), a large-scale \lt inferential knowledge dataset that contains 108K statements spanning four domains. We evaluate popular LLMs on \dataset; we find that state-of-the-art LLMs show significant performance drop (21\% relative drop for \gpt) on \lt data as compared to on head distribution data, and smaller models show even more generalization weakness. These results further underscore the necessity of \lt evaluation in developing generalizable LLMs.\footnote{\url{https://github.com/INK-USC/LINK}}

\end{abstract}

\begin{figure*}[ht]
\vspace{-0.5cm}
    \centering
    \captionsetup{width=\textwidth}
    \includegraphics[width=.75\textwidth]{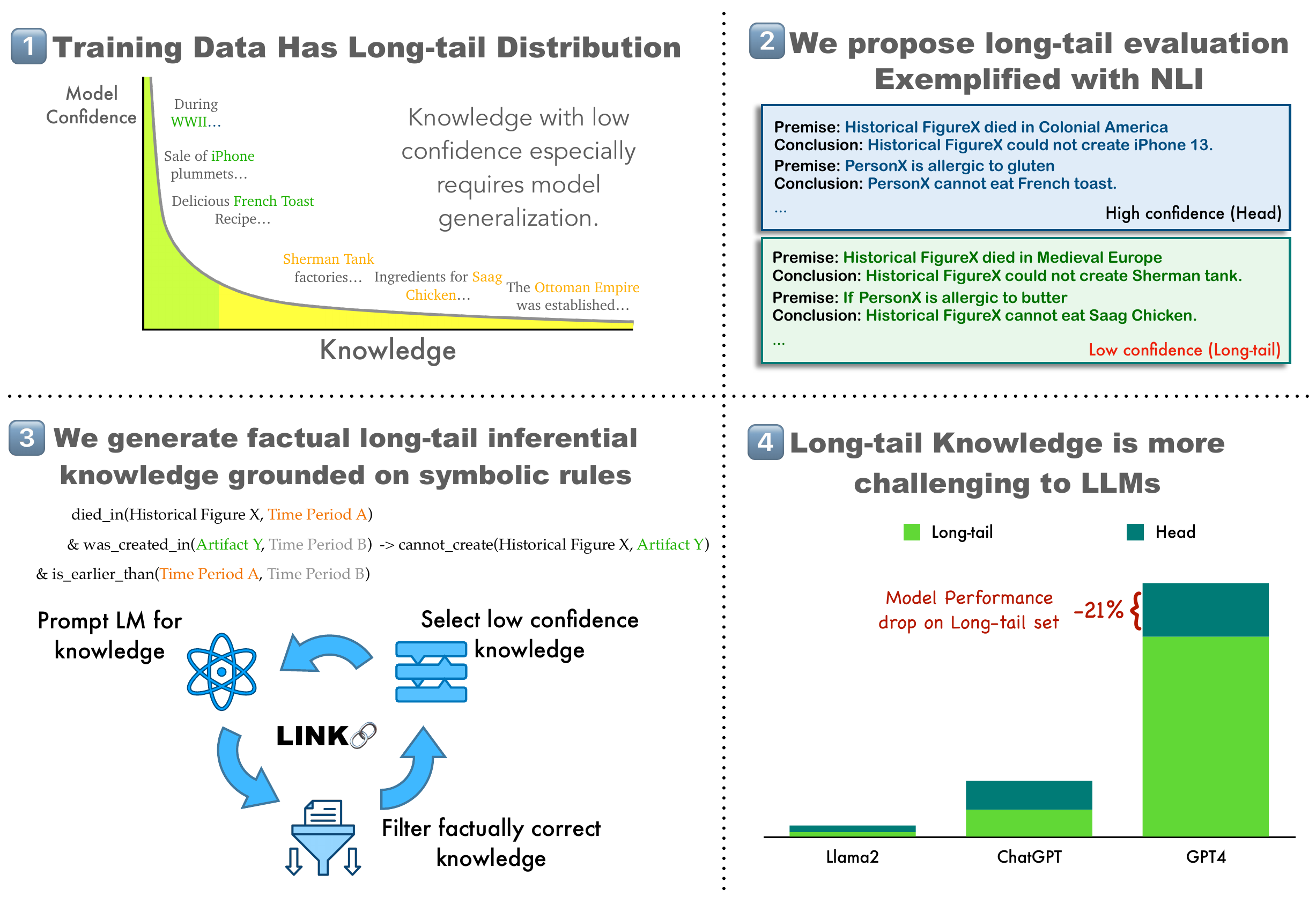}
    \caption{
        Overview of our motivation, \lt data generation framework, and model evaluation.
        }
        \looseness=-1
        
    \label{fig:teaser}
\end{figure*}
\section{Introduction}
\label{sec:introduction}
Generalization, especially to unfamiliar and novel situations, is a cornerstone for the usability of large language models (LLMs) in addressing varied real-world inquiries. 
This imminent demand necessitates evaluation of LLMs in the \ti{\lt} distribution (the space consisting of unfamiliar examples on which the model has low confidence). Previous works, mostly focusing on model memorization issue, define \lt knowledge using the frequency of entities in a knowledge base~\citep{cao2020open}, in the pre-training dataset~\citep{kandpal2023large}, or in Wikipedia search~\citep{mallen2022not}.
\citet{godbole2022benchmarking} introduces a general definition for \lt statements, where \lt examples are assigned \textit{lower likelihood by a pre-trained language model.} We follow this  definition which applies to various data format and language task -- for any set of statements with similar length and format, those in the \lt distribution cannot be generated or are generated with low confidence by the models, compared to those in the head distribution.
\looseness=-1

Recent works have noticed that LLMs have a marked decline in performance when facing inputs from the \lt~\cite{mccoy2023embers,razeghi2022impact}. Hallucination, for example, is found to be correlated with data being in the \lt distribution~\citep{li2024dawn,yu2024hallucidoctor}. LLMs' ineffective utilization of \lt knowledge impacts its reasoning capabilities and raises reliability concerns in downstream tasks~\citep{huang2023survey}.
\looseness=-1

Evaluation in the \lt distribution requires systematic generation of \lt data. However, obtaining examples in the \lt is non-trivial. With state-of-the-art LLMs being trained on vast volume of data on the internet~\citep{openai2023gpt,touvron2023llama2}, it is increasingly difficult to find unseen examples that can effectively test model generalization to its low-confidence end. Crowdsourcing \lt data is also difficult because of human cognitive bias~\citep{tversky1973availability,tversky1974judgment}, and LLMs' generation in the \lt distribution is hindered by their pretraining task of ``most likely"  next token~\citep{mccoy2023embers}.

While demonstrating \lt evaluation across all applications and domains is not feasible within the scope of one paper, this work focuses on inferential knowledge statement in the form of Natural Language Inference (NLI) task~\citep{bowman2015large,zellers2019hellaswag}: NLI requires extensive knowledge and complex reasoning about entities and events, and is one of such tasks on which LLMs have impressive performance~\citep{achiam2023gpt,touvron2023llama2,jiang2023mistral}. Following~\citet{sap2019atomic}, we structure inferential knowledge as \ti{if-then} relations with variables, written in a \textit{premise, conclusion} format (\Tabref{tab:examples}).

First, we make \lt inferential knowledge generation possible. We propose a novel and lightweight \lt inferential knowledge generation framework, \tf{L}ogic-\tf{In}duced-\tf{K}nowledge-Search (\tf{\framework}\benchmarkEmoji) (\Ssecref{sec:link}), a variable-wise prompting framework grounded on symbolic rules. This framework enables us to obtain \lt knowledge statements from existing LLMs. Our evaluation shows that by taking simply instructions \chat(\ttt{gpt-3.5-turbo}) and \gpt(\ttt{gpt-4}) can only produce statements in the head distribution that also have lower factual correctness, but using \framework~with the LLMs improves on both distribution conformity and factual correctness (\Ssecref{sec:generative}).



\begin{table}[h]
    \centering
    \scriptsize
    \caption{Examples of inferential knowledge in each domain of~\dataset, in a \textit{premise} (\textbf{P}), \textit{conclusion} (\textbf{C}) format.}
    \begin{tabular}{p{.1\textwidth}p{.05\textwidth}p{.7\textwidth}}
    
        \toprule
        \multirow{4}{*}{Locational} & Head & \makecell[tl]{
            \textbf{P:} Organization X has a branch in Great Lakes Region.\\
            \textbf{C:} Organization X has office in North America.
            } \\
        & \Lt & \makecell[tl]{
            \textbf{P:} Organization X has a branch in Tarapaca Region.\\
            \textbf{C:} Organization X has office in South America.
            } \\
        \midrule
        \multirow{4}{*}{
        \makecell[tc]{Outcome \\and Effect}} & Head & \makecell[tl]{
            \textbf{P:} Person X has Asthma.\\
            \textbf{C:} Person X should take Inhaled antiinflammatory drugs.
            } \\
        & \Lt & \makecell[tl]{
            \textbf{P:} Person X has Hepatitis.\\
            \textbf{C:} Person X should take Sofosbuvir.
            } \\
        \midrule
        \multirow{4}{*}{Temporal} & Head & \makecell[tl]{
            \textbf{P:} Plant X vanished in Paleolithic Era.\\
            \textbf{C:} Plant X cannot surround Notre Dame de Paris.\\
        }\\
        & \Lt & \makecell[tl]{
            \textbf{P:} Plant X vanished in Classical Greece.\\
            \textbf{C:} Plant X cannot surround Belém Tower.
            } \\
        \midrule
        \multirow{4}{*}{\makecell[tc]{Natural \\Properties}} & Head & \makecell[tl]{
            \textbf{P:} Bag X has trouble containing Clarinet.\\
            \textbf{C:} Upright Piano cannot fit in Bag X.\\
            } \\
        & \Lt & \makecell[tl]{
            \textbf{P:} Bag X has trouble containing Pandeiro.\\
            \textbf{C:} Dhak cannot fit in Bag X.\\
            } \\
        \bottomrule
    \end{tabular}
    \label{tab:examples}
    \vspace{-0.2cm}
\end{table}

Second, we test LLMs' \lt generalization capability on data generated by \framework (\Ssecref{sec:entailment}). We produce a large-scale dataset, \tf{L}ogic-\tf{In}duced-Long-\tf{T}ail (\tf{\dataset}), which contains 108k knowledge statements spanning across 4 different domains(\Tabref{tab:examples}). In the \lt test set of \dataset, \gpt's capability in identifying incorrect knowledge drop by 21\% relative to the test set in the head distribution, and the gap is even larger for other models we tested (\chat, \ttt{llama2-70b}). At the same time, human performance significantly outperforms LLMs in both distributions, and stays consistent between head and \lt test set.


Our work is the first to propose a systematic framework that generates data in the \lt distribution. Using NLI as an example, we show that generating data in the \lt distribution is an effective way for curating evaluation examples for LLM generalization. Our work serves as a starting point for the series of research on \lt data discovery and generation, and motivates the community to incorporate \lt evaluation into model building pipelines.
\looseness=-1

\begin{figure*}[ht]
\vspace{-0.5cm}
    \centering
    \includegraphics[width=0.8\columnwidth]{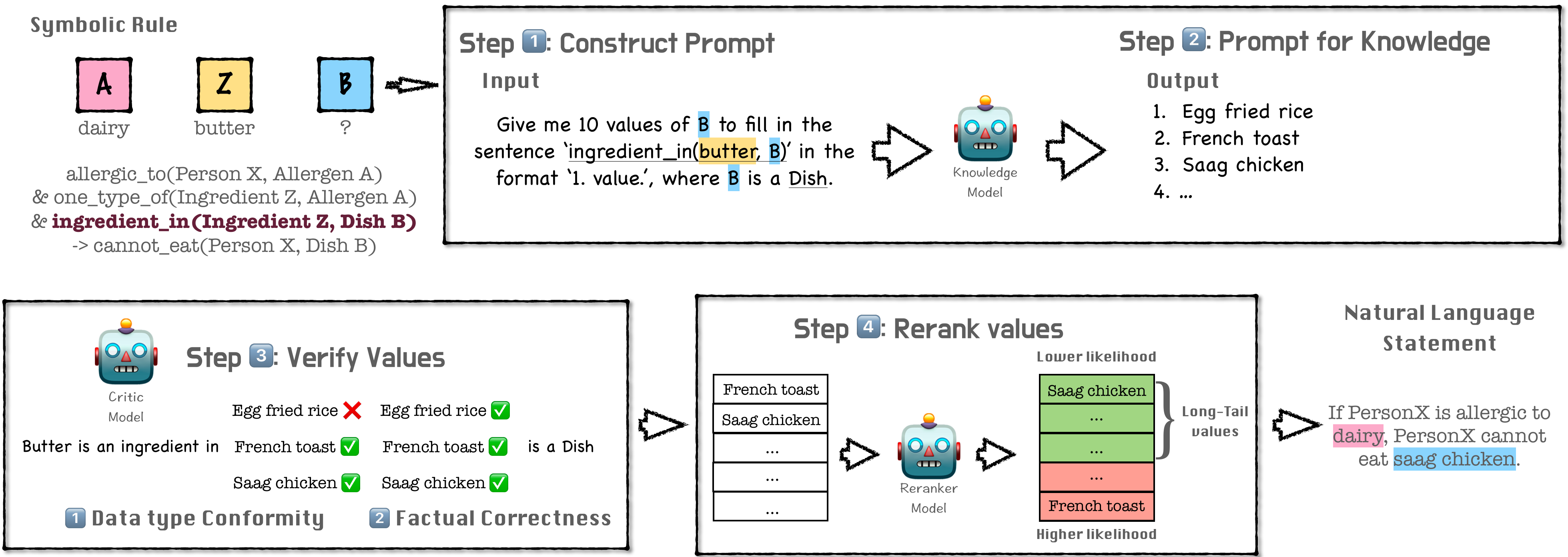}
    \caption{
    \looseness=-1
    Overview of knowledge beam search (\Ssecref{subsec:rule_verbalization}).
    We demonstrate searching \ti{B} conditioned on the values of \ti{A} and \ti{Z} from previous steps.  We only verbalize the predicates containing \ti{Person X} in the final statement as all other predicates contain knowledge that the model should have.}
    \label{fig:pipeline}
\end{figure*}

\section{\tf{L}ogic-\tf{In}duced-\tf{K}nowledge-Search (\tf{\framework}\benchmarkEmoji)}
\label{sec:link}
In this section, we first explain the advantages of generating inferential knowledge grounded on symbolic rules, then illustrate our process of curating symbolic rules, and lastly explain \textit{knowledge beam search}, our novel variable-wise search pipeline.
\looseness=-1

\subsection{Advantages of Generating Inferential Knowledge from Symbolic Rules}
Due to the fact that LLMs are pretrained with the task of generating the ``most likely"  next token, it is fundamentally challenging for them to directly generate \lt data through prompting that are factually correct and have low likelihood. Using symbolic rules to guide the generation of knowledge statements have three benefits: (1) Symbolic rules are \textit{designed to be correct}, so we alleviate the pressure of ensuring the deductive plausibility of the statement throughout the entire generation process. (2) The generation process can be broken down into multiple steps, each of which is conditioned on only one variable. Generating for one variable at a time will be much easier for the model, and it is easier to \textit{manipulate the distribution} of individual values than the entire sentence. (3) From one symbolic rule, one can get abundant combinations of variable values as long as they satisfy each predicate in the rule, making the generation process \textit{scalable}.


\subsection{Curating Symbolic Rules}
\label{subsec:rule_creation}
A symbolic rule consists of a premise and a conclusion. The conclusion is a single predicate, while the premise contains a set of predicates connected by $\&$ operators. Each predicate is a triple of a verb phrase, a subject and an object, and each variable in the symbolic rule has a designated data type.

While there are infinitely many ways to construct symbols rules, we create ours using the principles of \textbf{Compatibility and Mutual Exclusivity.} Compatibility refers to a scenario in which one or more events enables another event to occur. We construct the premise and conclusion such that if all predicates in the premise are true, then the predicate in the conclusion \textit{can} occur. Mutual Exclusivity refers to a scenario in which two or more events or conditions cannot occur simultaneously. We construct the premise and conclusion such that if all predicates in the premise are true, then the predicate in the conclusion \textit{will not} occur. To construct such conditions, we add constraints such as time, location, or outcomes to variables in the symbolic rules to make their interaction possible/desirable or impossible/undesirable. In other words, the conclusion describes an event (certain interaction between the variables) and the premise depicts some combination of conditions.

For example, a symbolic rule constrained by compatibility looks like:

\begin{table}[h]
\centering
\footnotesize
\begin{tabular}{l}
    
    \ti{exists(Person\;X, Location \;A)} 
    \ti{\& lives(Plant$\;$Y, Climate \;B)}\\
    \ti{\& has\_climate(Location \;A, Climate\;B)}\\
    $\to$ \ti{can\_plant(Person\;X, Plant\;Y)} 
\end{tabular}
\end{table}

And a symbolic rule constrained by mutual exclusivity looks like:

\begin{table}[h]
\centering
\footnotesize
\adjustbox{max width=\textwidth}{
\begin{tabular}{l}
    
    \ti{exists(Person\;X, Time Period \;A)} 
    \ti{\& exists(Plant$\;$Y, Time Period \;B)}\\
    \ti{\& is\_much\_later\_than(Time Period \;A, Time Period\;B)}\\
    $\to$ \ti{cannot\_plant(Person\;X, Plant\;Y)} 

\end{tabular}}
\end{table}



    


Here are some additional properties of our symbolic rules:



\paragraph{No tautologies.}
For example, \ti{allergic\_to(Person\;X, Allergen\;A)} $\to$ \ti{reacts\_badly\_to(Person\;X, Allergen\;A)} is not a valid symbolic rule. This is to ensure that the symbolic rule contains some reasoning.
    
\paragraph{The symbolic rule should contain at least 3 variables.} This is to ensure some degree of complexity in the symbolic rule.

\paragraph{The symbolic rule should not contain predicates out of scope of LLMs' knowledge.} \ti{has\_height(Tree\;X, Height\;Y)} is not a valid predicate, because it is unlikely that LLMs have knowledge about the exact height of one tree. This is to avoid hallucination.

We create symbolic rules that span across four domains (of constraint type): temporal, locational, outcome and effect, and natural properties, totaling 149 person-related rules and 268 object-related rules. More about symbolic rules in Appendix~\ref{app:rule_creation}.
\looseness=-1

\subsection{Knowledge Beam Search}
\label{subsec:rule_verbalization}

\paragraph{Defining search order.}
Since all variables are linearly chained, we can search them one by one without repetition. We always start with the subject of the sentence -- the person or the object, represented as \ti{Datatype X}. In the rule in~\Tabref{tab:baseline_prompt}, for example, we start with \ti{Person X} in the premise and find a chain of variables that connects it to the object in the conclusion: \ti{X, A, Z, B}.

For some rules that call for factual knowledge with only one correct answer, such as age, height, year, etc., we empirically find that it increases the knowledge quality to start from the subject in the conclusion and end with the object in the premise. 

\paragraph{Constructing Prompt.}
For each variable, we construct a prompt using all predicates that contain that variable and other previously searched variables. For example, to search variable \ti{B} in the rule in~\Tabref{tab:baseline_prompt}, we include predicate \ti{ingredient\_in(Ingredient\;Z, Dish\;B)}. We assume \ti{Z=butter} and construct the prompt as follows:

\ti{Give me 50 values of B to fill in the sentence ``ingredient\_in(butter, B)'' in the format ``1. value.'', where B is a Dish.}

\paragraph{Prompting for knowledge.}
For each partially searched beam, we obtain 200 values of the current variable from the knowledge model~\footnote{We use \ttt{text-davinci-003}~but one can use any model.}. We call OpenAI API 4 times, generating 50 values each time (temperature=0.7~\footnote{We keep \ttt{top\_p=1} for maximum diversity, and \ttt{top\_k} is unchangable. Ablation on temperature in Appendix~\ref{app:ablation-temperature}.}). After each call, we verify the values using a critic model (see paragraph below). To prevent duplicates, we explicitly instruct the model not to generate verified correct values and set \ttt{logit\_bias=-100} for incorrect values. We implement an early stop mechanism: if for two consecutive calls we do not get any correct values, we terminate the search for the beam.

\paragraph{Verifying values with a critic.}
We use \ttt{huggingface} default implementation of  \ttt{Flan-T5-XXL}~\citep{chung2022scaling}, an instruction-tuned model that can be used zero-shot, as the critic that checks data type conformity and factual correctness of the values.
We ask the model to output \ti{yes/no} on the correctness of a given statement. For data type conformity, the statement is ``\ti{\{value\} is a \{data type\}}.'' For factual correctness, we convert the symbolic predicate into a natural language statement. We obtain the \ti{yes} token probability and dynamically adjust the threshold for accepting values for different predicates. More about the critic model in Appendix~\ref{app:filter_job}.
\looseness=-1

\paragraph{Pushing values to \lt distribution with reranking.}
At each search step, we convert symbolic predicates into natural language statements (\ti{ingredient\_in(butter, saag chicken) $\to$ Butter is an ingredient in saag chicken}) and concatenate them with ``and''. We obtain the sentence likelihood using \ttt{huggingface} default implementation of \ttt{llama-7B}~\citep{touvron2023llama}\footnote{\ttt{llama2} was not released at the time of experiments.} and rerank the sentences from the \ti{lowest} likelihood to the \ti{highest} likelihood. We take top the 75\% values unless there are more than 200 values, in which case we take the top 200 values. Then we move on to the next variable. To control data distribution during evaluation, we also generate statements in the head distribution, by ranking the sentences from the \ti{highest} to the \ti{lowest} likelihood.

From 149 person-related rules and 268 object-related rules across four domains, we curate our dataset \tf{L}ogic-\tf{In}duced-Long-\tf{T}ail(\tf{\dataset}) that consists of 54K \lt knowledge statements. We also release 54K head distribution statements that are also searched with the \framework~framework. Domain-wise statistics of \dataset~in Appendix \ref{app:statistics}.
\looseness=-1

\section{Generating \Lt Inferential Knowledge with \framework}
\label{sec:generative}
In this section, we compare \framework's ability to generate \lt inferential knowledge with instruction-only LLMs, \chat~and \gpt, who do not use knowledge beam search.

\subsection{Instruction-Only Knowledge Generation}

\begin{table}[t]
    \begin{center}
    \scriptsize
    \begin{tabular}{p{.15\textwidth}p{.7\textwidth}}
        \toprule
        Symbolic Rule & \makecell[l]{
            \quad 
            is\_allergic\_to(Person X, Food allergen A)\& is\_\\ingredient\_in(Ingredient Z, Name of a dish or food B) \& \\ is\_one\_type\_of(Ingredient Z, Food allergen A)
            \\ $\to$ is\_not\_able\_to\_eat(Person P, Name of a dish or food B)
        }\\
        \midrule
        Prompt & \makecell[l]{
        In the following sentence, {\color{bittersweet}A} is a Food allergen, {\color{auburn} B} is a \\ Name of a dish or food, {\color{ao(english)} Z} is a Ingredient. Find values \\ of {\color{bittersweet}A}, {\color{auburn} B}, {\color{ao(english)} Z} to fill in the blank in the sentence `If Person \\ X is allergic to [{\color{bittersweet}A}] and [{\color{ao(english)} Z}] is a ingredient in [{\color{auburn} B}] and [{\color{ao(english)}Z}] \\ is one type of [{\color{bittersweet} A}], then Person X is not able to eat [{\color{auburn} B}].'\\ and make it a grammatical and correct sentence. \\Give me 50 values in the format `1. {\color{bittersweet}A}=, {\color{auburn} B}=, {\color{ao(english)} Z}='.
        }\\
        \bottomrule
    \end{tabular}
    \caption{An illustration of prompts for zero-shot LLMs, containing a symbolic rule in natural language and its variables with data type specified.}
    \label{tab:baseline_prompt}
    \end{center}
\end{table}
We generate knowledge statements from a subset of 200 symbolic rules from~\dataset~using \chat~and \gpt\ by only providing it with an instruction. We prompt the LLMs with a natural language version of the symbolic rule with data types of the variables specified, and ask them to populate the rule with all variables simultaneously(\Tabref{tab:baseline_prompt}). Generations from this prompt serve as the \textit{head distribution baseline}, to which we compare model generations when they are instructed to generate in the \lt distribution.

To instruct models to generate \lt knowledge from symbolic rules, we append \ti{``Use less frequent terms of A and B and C''} in the prompt.\footnote{We tried 10 prompts as shown in Appendix~\ref{app:longtail_prompts} and they have similar effect on model behavior.}

For each rule, we obtain 200 statements using default instruction and 200 statements using \lt instruction, from \chat~and \gpt~respectively. We present our findings below.

\subsection{\framework~Generations Consistently Fall in \Lt Distribution}
\label{subsec:distribution}
Following~\citet{godbole2022benchmarking}'s definition of \lt statements, we use the most capable LLM that was producing log likelihood at the time of experiments (\ttt{text-davinci-003}) to assign likelihood to generated data. We compare how different models assign distributions to the same statements, in order to ensure that the distribution stays consistent among all models despite the absolute log likelihood difference (Appendix~\ref{app:distribution_comparison}).
\looseness=-1


We calculate the log likelihood over \instruct~of all statements generated by \framework, compared with instruction-only \chat~and \gpt. We calculate $\delta = mean(D(H)) - mean(D(L))$ for each set of statements generated from each symbolic rule, where $D(\cdot)$ means the log likelihood distribution of the probability model, $H$ is the set of statements as head distribution baseline, and $L$ is the set of the statements intended to be in the \lt distribution. For a model to successfully generate \lt knowledge statements, the ``long-tail'' set of sentences should be assigned distinguishably lower probabilities than the sentence in the head distribution.

\begin{figure}[ht]
    \centering
    \includegraphics[width=\columnwidth]{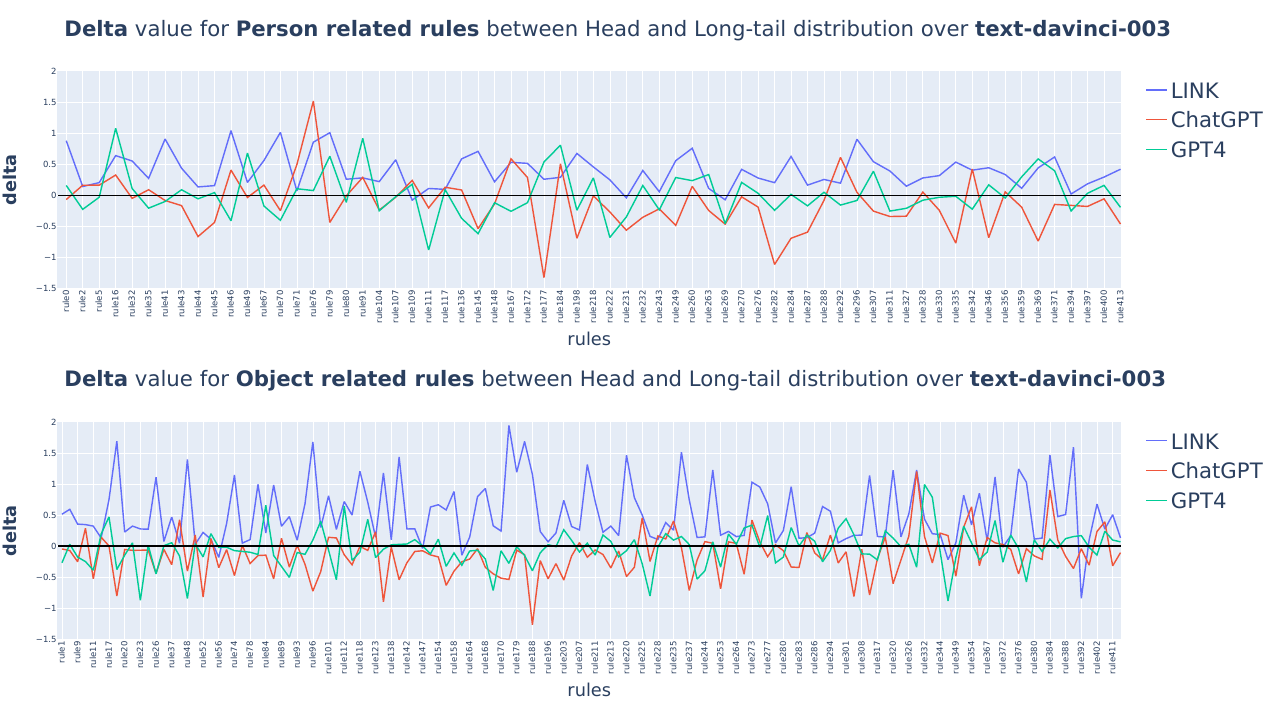}
    \caption{\framework~generations have higher $\delta$ values for most rules, while $\delta$ values of \chat~and \gpt~mostly locate around 0.}
    \label{fig:delta}
\end{figure}




\begin{figure}[ht]
    \captionsetup{width=\columnwidth}
    \includegraphics[width=\columnwidth]{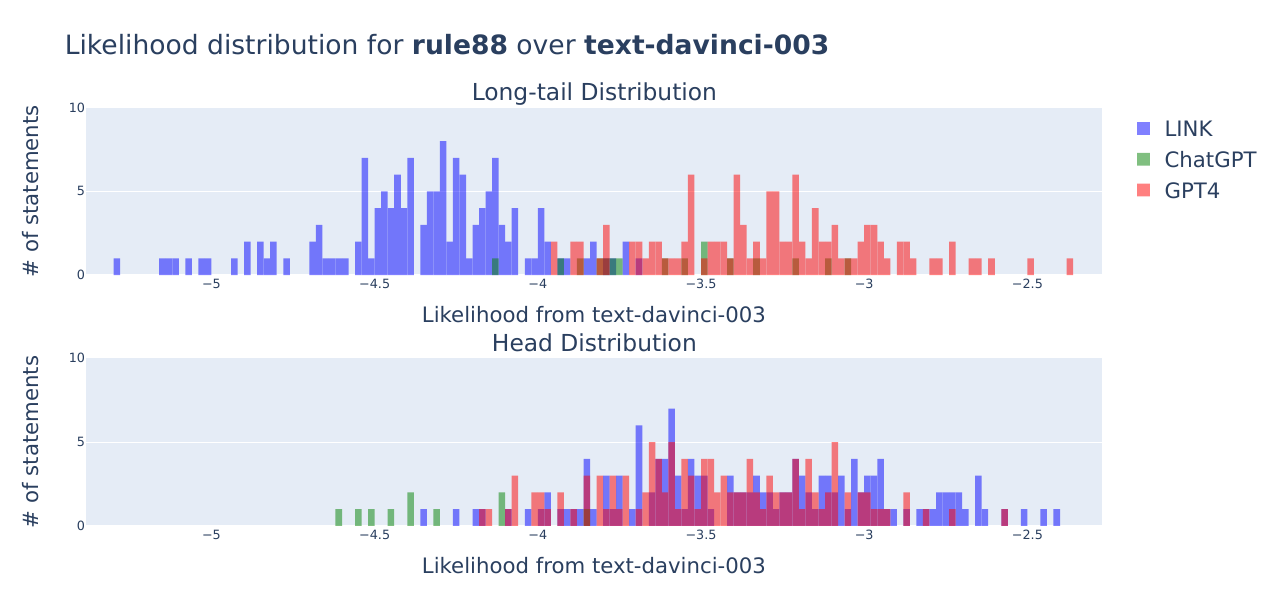}
    \caption{Only \framework~generations fall in the correct distributions on the log likelihood scale of \instruct.}
    \label{fig:distribution}
\end{figure}

\Figref{fig:delta} illustrates that while \chat\ and \gpt\ are not able to generate \lt statements merely from prompting, \framework~is able to generate \lt statements with much lower likelihood. 
Each grid on the x-axis represents a unique symbolic rule, and each grid on the y-axis represents $\delta \in [-1.5,2]$. A $\delta$ close to 0 means that the intended \lt distribution generations have the same log likelihoods as the statements in the head distribution, being larger than 0.3 empirically means a decent drop in likelihood, and being negative means the intended \lt distribution data have even high likelihood than the head distribution data.

Averaged across 200 sampled rules, \framework~has a positive $\delta$ of 0.48, while \chat~and \gpt~each has a delta of -0.14 and -0.02. The $\delta$ values for \framework~(blue line) float above 0 for most of the rules, some even being above 0.5. On the other hand, $\delta$ values for \chat~(red line) and \gpt~(green line) mostly locate around 0, with many being negative. 

To better illustrate the distribution of statements generated by \framework, \chat~and \gpt, we plot the log likelihood of the generated statements for one symbolic rule from the three methods~\Figref{fig:distribution}. To eliminate noise from incorrect statements on the distribution, we only plot the statements that are marked as correct in human evaluation (explained in \Ssecref{subsec:evaluation}). The likelihood distribution of more rules can be found in~\Figref{fig:multi}.

The \lt statements from \framework~clearly fall in a much lower probability distribution than \gpt's ``long-tail'' generations. Moreover, \gpt's generation in the ``\lt distribution'' in fact falls in the same probability distribution as its head distribution generations.


\subsection{\framework~Achieves Higher Data Correctness than Instruction-Only LLMs}
\label{subsec:evaluation}
\begin{table}[t]
        \centering
        \small
        \caption{\framework~has both the highest factual and data type accuracy in human evaluation.}
        \begin{tabular}{cccc}
        \toprule
          Accuracy & ChatGPT & GPT4 & \framework \\
           \midrule
             Data Type & 85.40 & 91.80 & \tf{94.23} \\
             Factuality & 67.50 & 84.82 & \tf{88.71} \\
             Overall & 56.44 & 78.23 & \tf{83.95} \\
        \bottomrule
        \end{tabular}
        \label{tab:evaluation}
\end{table}
In addition to distribution correctness, we also evaluate data type conformity and factual correctness of LLMs' \lt knowledge generations using crowdworkers from Amazon Mechanic Turk
(AMT). For data type conformity, we ask an AMT worker \ti{Is \{variable\} a \{data type\}?} for each variable in the symbolic rule. For factual correctness, we ask an AMT worker \ti{Does the premise entail the conclusion?} We sample 4,000 statements from \dataset~for human evaluation, of which 2,025 are from head distribution and 1,975 are from \lt distribution. We take 3 annotations for each statement and take the majority vote. Annotator agreement can be found in Appendix~\ref{app:agreement}. The AMT template can be found in Appendix~\ref{app:amt_template}.

\Tabref{tab:evaluation} shows that instruction-only \chat~and \gpt~underperform \framework~in both data type conformity and factual correctness.
Without \framework, both models struggle more with factual correctness, a foreseeable behavior in the low likelihood realm. For domain wise performance see~\Tabref{tab:human_eval_domain}. For examples of failure cases, see~\Tabref{tab:failure_cases}.

\subsection{Ablation Studies}
Our results above show that \framework\ is better than instruct-only generations with \chat\ and \gpt, in terms of both ``long-tailedness'' and factual correctness. In this section, we perform ablation studies on the role of reranker, critic, and knowledge models in \framework.

\paragraph{Ablation on reranker.}
Our variable-wise reranker is essential for pushing \framework\ generations into the \lt distribution.~\Figref{fig:ablation} presents the distribution comparison of generated statements by~\framework~and the variant without the reranker. Without the reranker, the statements for the \lt distribution are pulled towards the head distribution, making the two completely inseparable.
\looseness=-1

Post-hoc reranking over LLM generations does not have the same effect as variable-wise reranking in \framework.~\Figref{fig:posthoc} illustrates the distribution of generated statements by \framework, compared to instruction-only \gpt~and instruction-only \gpt~reranked by \instruct. Post-hoc reranking barely changes the distribution of generations, even when using the same model as the evaluation.
\looseness=-1

Even though log likelihood is both used by reranker and evaluation, they are taken from different models and using different inputs.
The statements we use for ranking the knowledge beams are shorter than the final statement, as they only consist of partial predicates. Despite these differences, variable-wise reranking that uses a smaller model achieves the separation that post-hoc filtering with the evaluation model cannot achieve.
\looseness=-1

Our findings highlight the importance of performing variable-wise reranking in \framework.

\paragraph{Ablation on the critic model.} 
Critic models are essential for guaranteeing the generation quality, especially in the \lt distribution. \Tabref{tab:ablation} in Appendix~\ref{app:ablation-critic} shows that removing the critic and removing both reranker and critic leads to significant drops of data type conformity and factual correctness of generations in the \lt distribution. Note that \framework~w/o reranker + critic has higher generation quality than \framework~w/o critic. This is because without a reranker the model is only able to generate statements in the head distribution, making it easier to be factually correct than in the \lt distribution. This observation further suggests that critic models are essential for generation qualities in the \lt distribution.

\paragraph{Ablation on Knowledge Model.}
Our analysis above has shown that by simply providing an instruction to \chat~and \gpt, we cannot effectively generate knowledge statements that are both high quality and in the \lt distribution. By adding \framework, we can improve both distribution correctness and generation quality, as shown in~\Figref{fig:link_gpt4} and~\Tabref{tab:link_gpt4}. 

Interestingly, we find that the generations from \framework~+ \gpt~do not show a big improvement over \framework~+ \instruct, both on distribution correctness and generation quality. This suggests that the improvement a stronger knowledge model brings to \framework~is marginal compared to that of the reranker and critic model. This finding highlights the effectiveness of \framework~in facilitating \lt generation regardless of the knowledge model.

\begin{table*}[t]
\vspace{-0.4cm}
    \centering
    \small
    \adjustbox{max width=\textwidth}{\begin{tabular}{cccccccccccc}
    \toprule
      Domain & Distribution  & \multicolumn{3}{c}{Llama2-70B} & \multicolumn{3}{c}{ChatGPT} & \multicolumn{3}{c}{GPT4} & \makecell[l]{Human \\Baseline}\\
          \midrule
           & & Pos & Neg & All & Pos & Neg & All & Pos & Neg & All & All \\
        \midrule          
         \multirow{3}{*}{\makecell[c]{Natural \\Properties}} & Head & 2.78 & 6.72 & \cellcolor{lightgray!25}0.44 & 1.45 & 5.56 & \cellcolor{lightgray!25}0.0 & 13.68 & 43.83 & \cellcolor{lightgray!25}9.23 & \cellcolor{lightgray!25}82.31 \\
            & \Lt & 2.64 & 3.49 & \cellcolor{lightgray!25}0.0 & 1.08 & 5.29 &\cellcolor{lightgray!25} 0.0 & 10.82 & 39.9 & \cellcolor{lightgray!25}7.21 & \cellcolor{lightgray!25}82.45 \\
            & $\Delta$ & - & - & \cellcolor{lightgray!25}-100\% & - & - & \cellcolor{lightgray!25}-0.0\% & - & - & \cellcolor{lightgray!25}-21.89\% & \cellcolor{lightgray!25} 0.17\%\\
         
         
         \midrule
         \multirow{3}{*}{Temporal} & Head & 6.58 & 2.6 & \cellcolor{lightgray!25} 0.46 & 10.72 & 38.28 & \cellcolor{lightgray!25} 4.13 & 68.3 & 43.95 & \cellcolor{lightgray!25} 36.91 & \cellcolor{lightgray!25} 84.69 \\
            & \Lt & 7.43 & 3.07 & \cellcolor{lightgray!25} 0.0 & 9.05 & 32.15 & \cellcolor{lightgray!25} 2.26 & 60.58 & 38.93 & \cellcolor{lightgray!25} 28.11 & \cellcolor{lightgray!25} 83.20\\
            & $\Delta$ & - & - & \cellcolor{lightgray!25} -100\% & - & - & \cellcolor{lightgray!25} -45.28\% & - & - & \cellcolor{lightgray!25} -23.84\% & \cellcolor{lightgray!25} -1.76\% \\

         
         \midrule
         \multirow{3}{*}{\makecell[c]{Outcomes \\ and Effects}} & Head & 7.62 & 7.93 & \cellcolor{lightgray!25} 1.22 & 19.92 & 29.27 & \cellcolor{lightgray!25} 6.1 & 57.32 & 55.18 & \cellcolor{lightgray!25} 41.16 & \cellcolor{lightgray!25} 83.83 \\
          & \Lt & 8.5 & 9.97 & \cellcolor{lightgray!25} 0.59 & 19.65 & 26.1 & \cellcolor{lightgray!25} 2.64 & 55.72 & 46.63 & \cellcolor{lightgray!25} 33.43 & \cellcolor{lightgray!25} 85.13 \\
          & $\Delta$ & - & - & \cellcolor{lightgray!25} -51.64\% & - & - & \cellcolor{lightgray!25} -56.72\% & - & - & \cellcolor{lightgray!25} -18.78\% & \cellcolor{lightgray!25} 1.56\% \\
         
         
         \midrule
         \multirow{3}{*}{Locational} & Head & 8.57 & 7.14 & \cellcolor{lightgray!25} 0.0 & 17.14 & 10.0 & \cellcolor{lightgray!25} 5.71 & 18.57 & 18.57 & \cellcolor{lightgray!25} 2.86 & \cellcolor{lightgray!25} 75.71 \\
            & \Lt & 11.24 & 8.99 & \cellcolor{lightgray!25} 3.37 & 15.73 & 4.49 & \cellcolor{lightgray!25} 1.12 & 37.08 & 8.99 & \cellcolor{lightgray!25} 3.37 & \cellcolor{lightgray!25} 67.42\\
            & $\Delta$ & - & - & \cellcolor{lightgray!25} 0.0\% & - & - & \cellcolor{lightgray!25} -80.39\% & - & - & \cellcolor{lightgray!25} 17.83\% & \cellcolor{lightgray!25} -10.95\%\\
         
         
         \midrule
         \multirow{3}{*}{Total} & Head & 5.08 & 5.28 & \cellcolor{lightgray!25} 0.56 & 8.21 & 20.67 & \cellcolor{lightgray!25} 2.62 & 39.49 & 44.87 & \cellcolor{lightgray!25} 23.64 & \cellcolor{lightgray!25} 83.12 \\
            & \Lt & 5.69 & 4.67 & \cellcolor{lightgray!25} 0.27 & 7.76 & 17.86 & \cellcolor{lightgray!25} 1.28 & 36.58 & 39.34 & \cellcolor{lightgray!25} 18.66 & \cellcolor{lightgray!25} 82.44 \\
            & $\Delta$ & - & - & \cellcolor{lightgray!25} \textbf{-51.79\%} & - & - & \cellcolor{lightgray!25} \textbf{-51.15\%} & - & - & \cellcolor{lightgray!25} \textbf{-21.07}\% & \cellcolor{lightgray!25} \textbf{-0.82\%} \\
         
         
    \bottomrule
    \end{tabular}}
    \caption{Performance on the entailment classification task of three LLMs decreases on the \lt distribution compared to the head distribution, while human performance does not.} 
    \label{tab:probing}
\end{table*}

\section{LLMs' (Lack of) Generalization in the \Lt Distribution}
\label{sec:entailment}

Using data from \dataset, we evaluate LLM generalization through an \textit{entailment classification} task on inferential knowledge in the \lt distribution.

We use all human evaluated knowledge statements except for those with incorrect data types in \dataset, with 1,925 statements in head distribution and 1,856 statements in \lt distribution. Statements rated as factually correct has entailment between premise and conclusion, and statements rated as factually incorrect has contradiction between premise and conclusion.



In order to prevent LLMs' template biases from misleading the evaluation, we convert each statement into 13 question templates, where each question templates corresponds to a positive label (\textit{``Yes'', ``True'', or ``Right''}) or a negative label (\textit{``No'', ``False'', or ``Wrong''}). The question templates are summarized in Appendix~\ref{app:probing_template}. We consider a model answering accurately about one statement \textit{only if} the model answers \textit{all} question templates correctly.

We evaluate three LLMs: \ttt{llama2-70B}, \chat~and \gpt. In order to enforce the model to predict the target token sets and minimize format noncompliance, we use Chain-of-Thought (CoT)~\citep{wei2022chain} prompting that includes 2 in-context examples with randomly shuffled orders of positive label and negative label.


For each domain, we report aggregated (All) performance of each model as well as human baseline performance in~\Tabref{tab:probing}.
We also include performance on positive labels only and negative labels only. We also mark relative performance drop $\delta = \frac{t-h}{h}$, where $h$ and $t$ are head and tail distribution aggregated performance. 
\looseness=-1

We obtain human performance on the same set of statements. We recruit 17 AMT workers who do not participate in the evaluation task (and thus have not seen the task data). The workers see the knowledge statements in \ti{premise, conclusion} format and are asked to select ``\ti{yes/no}'' to whether the premise entails the conclusion. The workers are asked to use search engines to verify their answers. See AMT templates in Appendix~\ref{app:amt_template}.

We make the following observations on LLM generalization in \lt NLI.




\paragraph{Performance drops in the \lt distribution.}
All models exhibit a large relative drop in performance in the \lt distribution. The most competitive model, \gpt, has a 21\% overall drop from head to \lt distribution, while other models exhibit a even larger drop.

\paragraph{Human Performance does not drop for \lt distribution.}
Performance drop in the \lt distribution does not occur to humans for 3 out of 4 domains. It is expected because humans can verify their knowledge using search engines, so infrequent knowledge does not challenge humans as much as models (discussion in Appendix~\ref{app:discussion}). The exception with the locational domain may be due to some relations being less available online(eg. \ti{banned\_in(Food, Country)}). 
\looseness=-1

\paragraph{Brittleness towards question templates.}
The huge gap between model and human baseline performance indicate that LLMs cannot reliably reason on the same statement when question templates change. We find that model performance between positive and negative labels can be very different for certain domains, indicating that models are miscalibrated for positive and negative answer tokens. Although this phenomenon is not entirely due to the shift of distribution of the knowledge statements, it is caused by model's unfamiliarity of certain question templates. For example, we find that template 5 and template 12 are the same question (\textit{``Premise: ... Conclusion: ... Does the Premise entail the Conclusion?"})  with opposite answers (``Yes" and ``No"), but all models' performance on template 5 is significantly lower than that on template 12. This suggests that models are more likely to create false negatives in such context, another evidence of performance drop due to \lt distribution.

In summary, our analysis shows that while human's inferential reasoning is not affected by their familiarity to the data (provided that they know the entities involved), model's inferential reasoning ability drops over \lt knowledge. Our result highlights the importance and effectiveness of \lt evaluation for model generalization, and our dataset \dataset~can be used as a useful resource for testing inferential generalization of LLMs.

\section{Related Work}


Works on model generalization analysis have focused on \textbf{generating adversarial examples} for model evaluation~\citep{zhang2019adversarial,ziegler2022adversarial,perez2022red,casper2023explore}, flagging abnormal inputs that are likely to trigger bad behavior. Recently, the community has realized the importance of \tf{testing language models' abilities in the \lt distribution}~\citep{godbole2022benchmarking}. Works reveal that LLM performance is affected by input data probability.~\citep{mccoy2023embers, razeghi2022impact}, and more works have focused on \tf{generating less common data for probing LLMs}. RICA~\citep{zhou2020rica} proposes to include novel entities in self-contained commonsense statements to evaluate robust inference capabilities. UnCommonSense~\citep{arnaout2022uncommonsense} proposes to evaluate models on informative negative knowledge about everyday concepts in addition to positively expressed commonsense knowledge.~\citet{razeghi2022impact} observe a correlation between the model performance on math problems and the frequency of numeric and temporal terms from those instances in the pretraining data.

\looseness=-1
In addition to probing models on less common data, recent works also \tf{test LLMs generating less common data}.~\citet{chen2023say} propose a negative knowledge generation task where models generate uncommon knowledge with negation conditioned on constrained keywords.~\citet{tang2023less} introduce the “less likely brainstorming” task that asks a model to generate outputs that humans think are relevant but less likely to happen.
 
Generating uncommon data is challenging not only for LLMs, but also for \tf{humans because of our cognitive bias}.~\citet{tversky1974judgment} observe that humans are prone to more systematic errors when facing uncertain events, and~\citet{tversky1973availability} reveal that humans tend to evaluate the frequency of classes or the probability of events by availability, i.e., by the ease with which relevant instances come to mind. These traits make it difficult for humans to come up with novel associations~\citep{kray2006thinking}, a crucial ability to create data in the \lt distribution.
\section{Conclusion}
\label{sec:conclusion}
Using NLI as a case study, we illustrated the significant potential of \lt data in uncovering the generalization limitations of LLMs. We introduced the first systematic framework designed to generate inferential data within the \lt distribution, and then demonstrated a noteworthy performance drop of LLMs in the \lt examples. Our work initiates a new line of research focused on \lt data discovery and generation, urging the research community to adopt \lt evaluation in the development of LLMs.

\newpage

\newpage
\section*{Checklist}
\subsection*{Limitation}
\paragraph{Limitation on knowledge statement format.} \Lt knowledge statements may come in multiple shapes and forms. Our work focuses only on \ti{premise, conclusion} format, as the first step towards the generation of knowledge statements. The symbolic rules do not have high complexity, due to the limited number of variables and predicates, and being under the constraint for the symbolic rules to be linearly chained. Therefore, the effectiveness of our framework on generating more complex knowledge statements has not been tested. 

\paragraph{Limitation on testing with open-source models.} Our work did not include open-source models in evaluations of \lt statement generation and entailment classification task. While \chat~and \gpt~are arguably the strongest models, open-source models may exhibit new behaviors in the \lt realm that are worth exploring.

\paragraph{Limitation on ablating with different critic and reranker model settings.} While we performed extensive ablation studies on the critic and reranker models and established their importance in the~\framework~framework, we did not explore a diverse set of model options as well as hyperparameter settings. Using other models may or may not affect the performance of~\framework.

\paragraph{Limitation on sample size.} Due to constraint from human annotation resources, we were only able to evaluate models on 200 rules uniformly sampled from the \dataset. Although the general trend should remain the same, model performance evaluated on all rules may result in some deltas.

\subsection*{Risk}

\paragraph{Generation of harmful values.} \framework~might be used on mal-intention-ed rules or searching for toxic and harmful values, where researchers may replace our reranker with another model trained to prefer more harmful values.

\paragraph{Environmental tax.} Another potential risk is increasing environmental burdens because we extensitvely call OpenAI APIs to large language models during search; however, one can replace the large language models with smaller open source models with less environmental tax.

\paragraph{Factual errors in generations.} Because~\framework~operates in the \lt realm, its generations are no guaranteed to be correct 100\% of the time. If one uses the generations directly without verification, one my introduce false information into their system.

\subsection*{Use and Distribution}

All data we collected through LLMs in our work are released publicly for usage and have been duly scrutinized by the authors. Data for all human studies that we conduct are also publicly released with this work, with appropriate annotator anonymizations.

Our framework \framework~may only be used for generations that follow the ethics guideline of the community. Using \framework~on mal-intention-ed rules or searching for toxic and harmful values is a potential threat, but the authors strongly condemn doing so.

\bibliography{custom,anthology}
\newpage
\appendix


\section{More Discussions}
\label{app:discussion}
Our main experiments on entailment classification task assume that the LLMs have the specific knowledge necessary to answer the questions. While this is a fair implication, given that their training data are known to contain essentially anything available on the web, one may also worry that the model's handling of such knowledge is impacted by the memorization of this knowledge. As knowledge gets more \lt, it means it appears in the training data less and is thus harder to memorize. Given the imperfect memorization of current LLMs, this may impact their performance on such knowledge, and our current experiments suggest that. To verify this, for future experiments one can count the frequency of this \lt knowledge in the training data to measure how imperfect memorization makes certain knowledge \lt.

\section{Symbolic Rule Creation}
\label{app:rule_creation}

Following the criteria mentioned in~\Ssecref{subsec:rule_creation}, we curated 417 symbolic rules using the following steps:

\begin{enumerate}
    \item Select Compatibility (conclusion will be positive) or Mutual Exclusivity (conclusion will be negative).
    \item Defining Constraints. Our constraints can be categorized as temporal, locational, natural properties, and desirable/undesirable outcomes and effects. \textit{Temporal} constraints refer to time period or age, \textit{locational} constraints refer to geographic location such as countries or cities as well as climates (tropical or polar, etc), \textit{natural properties} refer to physical properties of objects such as temperature, size, density, speed, and \textit{outcomes and effects} include allergies, cure of disease, etc.
    \item Selecting argument types. We either use Person-Object or Object-Object as key arguments.
    \item Defining interactions between arguments. Person-Object interactions include ``using'', ``operating'', ``buying'', ``consuming'', and Object-Object are more type-specific (eg. ``scratching'' when the constraint is ``hardness'' in natural property).
    \item Expanding data types for Person or Object. We prompt \instruct~to generate more specific data types under Person (eg. Historical Figure) or Object (eg. Vehicle, Tool).
    \item Optimize the verb describing the interaction. We prompt \instruct~to generate a more accurate verb for the expanded data types.
\end{enumerate}
\Tabref{tab:rule_expansion_prompt} shows how an example symbolic rule is constructed.
\begin{table}[h]
    \centering
    \small
    \caption{Illustration of our process for creating an example symbolic rule.}
    \begin{tabular}{p{.2\textwidth}p{.7\textwidth}}
        \toprule
        Constraint & Temporal (Age) \\
        \midrule
        Arguments & Person-Object \\
        \midrule
        Constraint Predicate & \makecell[tl]{
            is\_of\_age(Person A, Age X) \&  requires\_a\_\\minimal\_age\_of(Object B, Age Y) \\ \& is\_smaller\_than(Age X, Age Y)\\
            } \\
        \midrule
        Principle & Mutual Exclusivity \\
        \midrule
        Interaction & cannot\_\textbf{operate}(Person A, Object B) \\
        \midrule
        Prompt for Data Type Expansion & \makecell[tl]{In rule ``requires\_a\_minimal\_age\_of(Object \\ B, Age Y) \& 
        cannot\_operate(Person A, \\ Object B)", B is a variable representing an \\ object. List 10 subcategories of object that \\ B could be that also make the rule true.}\\
        \midrule
        Expanded Data Types & \makecell[tl]{Vehicle, Machinery, Alcohol, Firearm, \\ Tattoo Equipment, Tobacco Product} \\
        \midrule
        Prompt for Verb Optimization & \makecell[tl]{ cannot\_operate(Person A, Object B) is equal \\ to [mask](Person A, Vehicle B).
        Write the \\ best predicate that could fit in [mask] token.} \\
        \midrule
        Expanded Rule Example & \makecell[tl]{
        is\_of\_age(Person A, Age X) \& requires\_a\_\\minimal\_operating\_age\_of(Object B, Age \\ Y) \& is\_smaller\_than(Age X, Age Y)} $\to$ cannot\_drive(Person A, Vehicle B)\\
        \bottomrule
    \end{tabular}
    \label{tab:rule_expansion_prompt}
\end{table}

\section{Critic Model}
\label{app:filter_job}
We find that while the critic model usually verifies data type conformity with high accuracy, it often creates false negatives when verifying factual correctness. Moreover, even within false negatives that result from the same predicate, the correct values get higher \ttt{yes} token probabilities than the incorrect values. We hypothesize that while the critic model is less confident about certain knowledge because it is trained on a smaller portion of the knowledge than \ttt{text-davinvi-003}, it can still rank the values inherently. Therefore, we extract the probability of the \ttt{yes} token instead of taking the argmax. We also implement a dynamic critic threshold that adjusts the threshold for accepting values for different predicates. The algorithm is as follows:

\begin{enumerate}
    \item We start with a threshold of 0.85.
    \item If no correct values are found, we decrease the threshold by 0.05.
    \item If some correct values are found, we set the threshold for the predicate to the current threshold and do not decrease it in further calls.
    \item If the threshold is set but we find some values with a higher \ttt{yes} token probability than the threshold, we increase the threshold by an increment of 0.05 to accommodate the higher probability. Then we retrospectively reject previous accepted values with a lower \ttt{yes} token probability than the new threshold.
    \item For data type conformity, we set a minimum threshold of 0.65 because we expect the model to be more confident.
\end{enumerate}

In this way, we can find the maximum available threshold for each beam, which guarantees precision while reducing false negatives.

To verify the effectiveness of our critic model, we use crowd workers from AMT to evaluate the data type conformity and factual correctness of predicates. Specifically, for each symbolic predicate that contains two variables (e.g., \ti{exist\_during(Location X, Historical Time Period Y)}), we will present a statement in natural language (e.g., \ti{Saigon} existed during \ti{The Cold War}.) with 3 types of questions: (1) clear reference: Q1 and Q2. (2) factual correctness: Q3. (3) data type conformity: Q4 and Q5.

\begin{itemize}
    \item \textbf{Q1: } Does ``\ti{Value A}'' in the Statement ``\ti{Statement}'' have a clear reference?
    \item \textbf{Q2: } Does ``\ti{Value B}'' in the Statement ``\ti{Statement}'' have a clear reference?
    \item \textbf{Q3: } Is the Statement ``\ti{Statement}'' factually correct, with very high probability?
    \item \textbf{Q4: } Is the Statement ``\ti{Value A} is a \ti{Data Type A}.'' factually correct, with very high probability?
    \item \textbf{Q5: } Is the Statement ``\ti{Value B} is a \ti{Data Type B}.'' factually correct, with very high probability?
\end{itemize}

We sample 3 rules from our data and requested human annotators to rate the data type conformity and factual correctness of statements. \Tabref{tab:verifier} shows the error rate of each question. Only if all the questions are answered with ``Yes'' do we consider the statement as correct. The overall correctness of statements in head distribution and \lt distribution are 0.8567 and 0.8467 respectively, which indicates a high quality of statements accepted by our critic model.
\begin{table}[ht]
    \centering
    \small
    \begin{tabular}{cccccc}
        \toprule
        & Q1 & Q2 & Q3 & Q4 & Q5\\
        \midrule
        Error Rate & 0.0004 & 0 & 0.0639 & 0.0011 & 0 \\
        \bottomrule
    \end{tabular}
    \caption{The error rate of each question in human verification. Most errors occur on factual correctness.}
    \label{tab:verifier}
\end{table}



\section{Ablation Studies on~\framework}

\subsection{Hyperparameters on Knowledge Model}
\label{app:ablation-temperature}
When constructing \dataset, we used \instruct~as the knowledge model with temperature=0.7 and top\_p=1. Since top\_p=1 maximizes sampling diversity and top\_k is hidden from the OpenAI API, we conduct ablation studies on whether temperature affects the result of knowledge search, comparing temperature of 0.5 (low diversity), 0.7 (medium diversity) and 1.0 (high diversity), using a few sampled rules. In this ablation study, we use \ttt{gpt-3.5-turbo-instruct} checkpoint as knowledge model and \ttt{llama-2-70b} as the approximation of the language distribution.

\begin{table}[h]
    \centering
    \small
    \caption{Different temperatures result in similar data type conformity and factual correctness.}
    \begin{tabular}{cccc}
    \toprule
      Temperature & Data Type & Factuality & Overall \\
       \midrule
       0.5 & 89.05 & 93.21 & 82.94 \\
       0.7 & 89.00 & 92.33 & 82.25 \\
       1.0 & 88.08 & 91.75 & 81.00 \\
    \bottomrule
    \end{tabular}
    \label{tab:hyperparam}
\end{table}

\Tabref{tab:hyperparam} shows similar data type conformity and factual correctness among the three ablated temperature, with temperature=1.0 having the lowest accuracy among the three settings.

\begin{figure*}
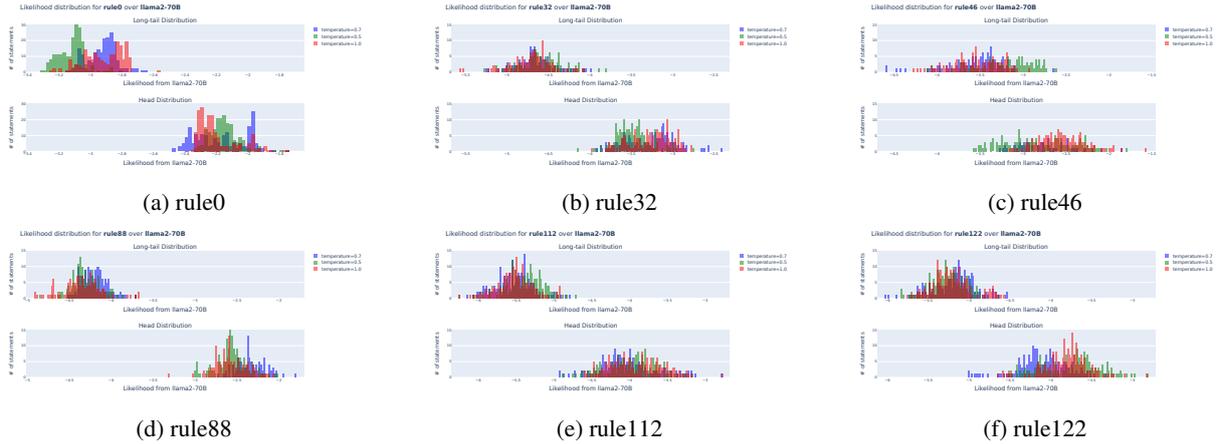

    \centering
    \begin{subfigure}{.3\linewidth}
        \centering
        \includegraphics[width=\textwidth]{images/hyperparam/rule0.pdf}
        \caption{rule0}\label{fig:image11}
    \end{subfigure}
        \hfill
    \begin{subfigure}{.3\linewidth}
        \centering
        \includegraphics[width=\textwidth]{images/hyperparam/rule32.pdf}
        \caption{rule32}\label{fig:image12}
    \end{subfigure}
        \hfill
    \begin{subfigure}{.3\linewidth}
        \centering
        \includegraphics[width=\textwidth]{images/hyperparam/rule46.pdf}
        \caption{rule46}\label{fig:image13}
    \end{subfigure}
    
    \begin{subfigure}{.3\linewidth}
        \centering
        \includegraphics[width=\textwidth]{images/hyperparam/rule88.pdf}
        \caption{rule88}\label{fig:image21}
    \end{subfigure}
        \hfill
    \begin{subfigure}{.3\linewidth}
        \centering
        \includegraphics[width=\textwidth]{images/hyperparam/rule112.pdf}
        \caption{rule112}\label{fig:image22}
    \end{subfigure}
        \hfill
    \begin{subfigure}{.3\linewidth}
        \centering
        \includegraphics[width=\textwidth]{images/hyperparam/rule122.pdf}
        \caption{rule122}\label{fig:image23}
    \end{subfigure}
    \caption{All three temperature settings of \framework~can successfully generate knowledge statements in the \lt~distribution, except for when temperature=0.5 in one of the six sampled rules.}
    \label{fig:hyperparam}
\end{figure*}

\Figref{fig:hyperparam} shows that all three temperature settings can successfully generate knowledge statements in the \lt distribution, except for when temperature=0.5 in one of the six sampled rules.

This phenomenon reflects that higher temperature helps generating more diverse values and therefore more likely to generate \lt values, while risking lowering factual salience.

\subsection{Effect of Critic on~\framework}
\label{app:ablation-critic}
To investigate the effectiveness of the critic, we provide an ablation study on a few sampled rules by removing the critic in~\framework. \Tabref{tab:ablation} shows the generation quality of \framework~and several variants in \lt distribution. Without the critic, the generation quality decreases significantly. However, the performance drop is less significant in the head distribution~\Tabref{tab:ablation-head}. Besides, if we replace the reranker with a random sampling method, the generated statements cannot lie in the \lt distribution (which will be further explained in \Ssecref{subsec:ablation-reranker}) and have higher correctness without the critic. It indicates that it is harder for models to generate correct statements from the \lt distribution than the head distribution without \framework.

    \begin{table}[h]
        \centering
        \small
        \caption{Ablation study on the critic model in the \lt distribution. Removing the critic from \framework~will significantly decrease the generation quality. Using a critic is necessary to guarantee the correctness of generated statements, especially in the \lt distribution.}
        \begin{tabular}{lccc}
        \toprule
           & Data Type & Factuality & Overall \\
           \midrule
           \framework & \textbf{93.42} & \textbf{97.50} & \textbf{91.33} \\
           \ \ w/o critic & 52.58 & 52.08 & 33.00 \\
           \ \ w/o critic+reranker & 75.42 & 73.92 & 58.25\\
             \midrule
           \framework~with~\gpt & 92.75 & 96.17 & 89.25 \\
           \ \ w/o critic & 63.00 & 58.17 & 40.00 \\
           \ \ w/o critic+reranker & 88.33 & 83.50 & 74.50\\
        \bottomrule
        \end{tabular}
        \label{tab:ablation}
    \end{table}

    \begin{table}[h]
        \centering
        \small
        \caption{Ablation study on the critic model in the head distribution. Removing the critic decreases the data quality, but not as much as in the \lt distribution. \framework~w/o critic+reranker has the same performance between head and \lt distribution, demonstrating that without a reranker all generations are in the same distribution.}
        \begin{tabular}{lccc}
        \toprule
           & Data Type & Factuality & Overall \\
           \midrule
           \framework & \textbf{95.17} & \textbf{97.75} & \textbf{93.00} \\
           \ \ w/o critic & 80.33 & 73.50 & 59.75 \\
           \ \ w/o critic+reranker & 76.66 & 74.83 & 59.33\\
             \midrule
           \framework~with \gpt~& 92.17 & 98.08 & 90.83 \\
           \ \ w/o critic & 91.00 & 80.42 & 72.08 \\
           \ \ w/o critic+reranker & 88.17 & 85.33 & 75.58\\
        \bottomrule
        \end{tabular}
        \label{tab:ablation-head}
    \end{table}

\subsection{Effect of Reranker on~\framework}
\label{subsec:ablation-reranker}
\begin{figure*}
    \centering
    \begin{subfigure}{.3\linewidth}
        \centering
        \includegraphics[width=\textwidth]{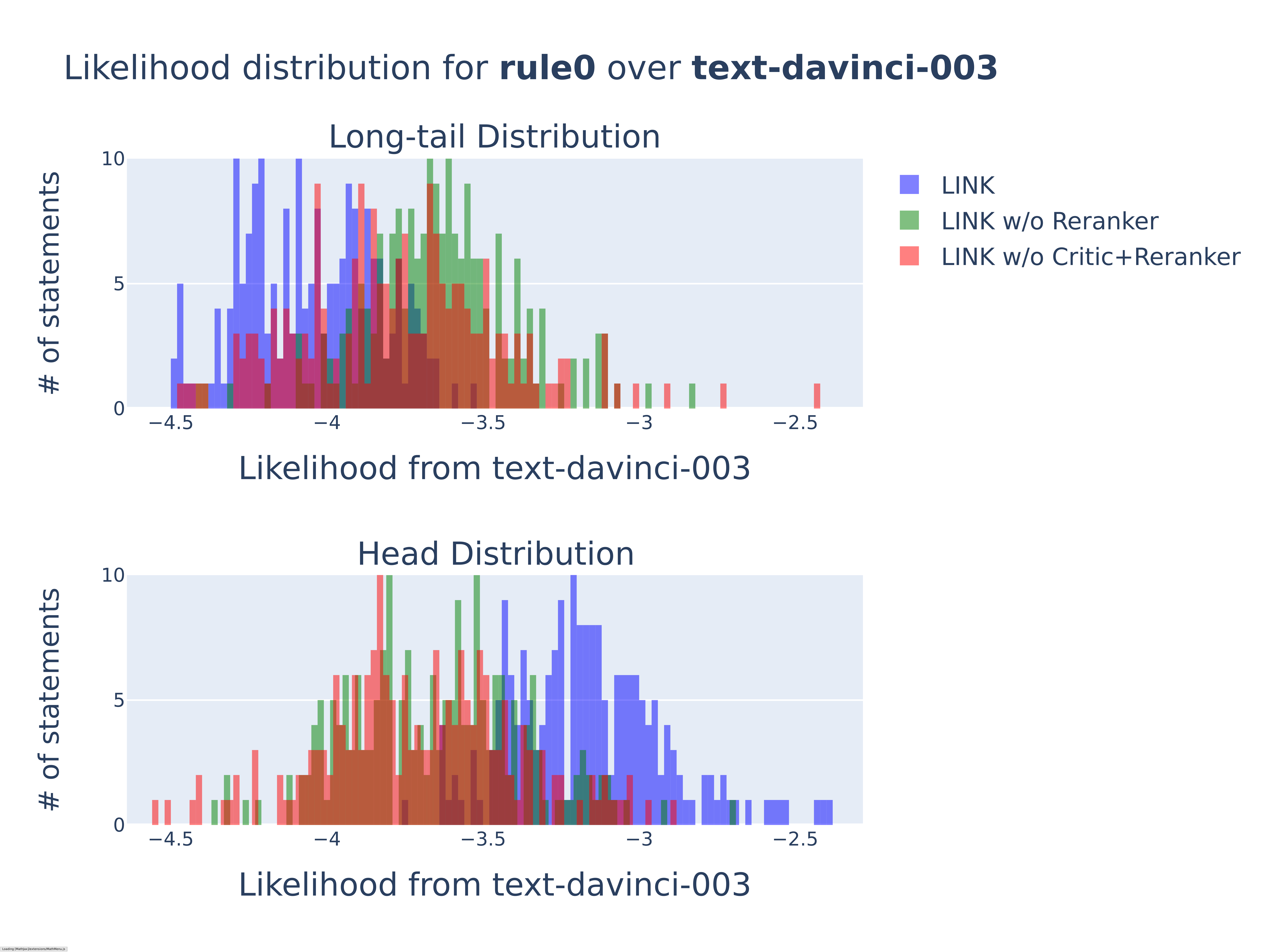}
        \caption{rule0}
    \end{subfigure}
        \hfill
    \begin{subfigure}{.3\linewidth}
        \centering
        \includegraphics[width=\textwidth]{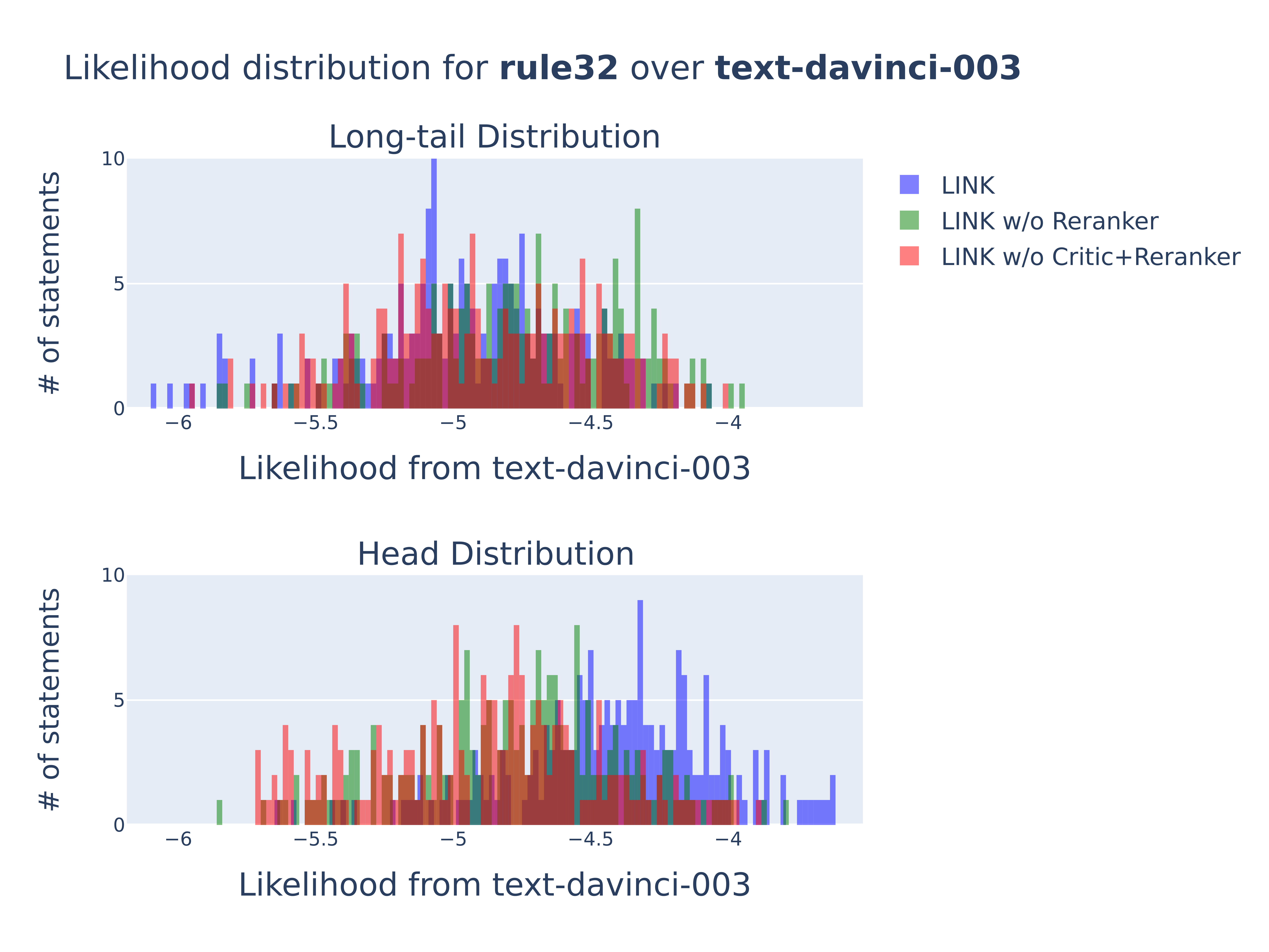}
        \caption{rule32}
    \end{subfigure}
        \hfill
    \begin{subfigure}{.3\linewidth}
        \centering
        \includegraphics[width=\textwidth]{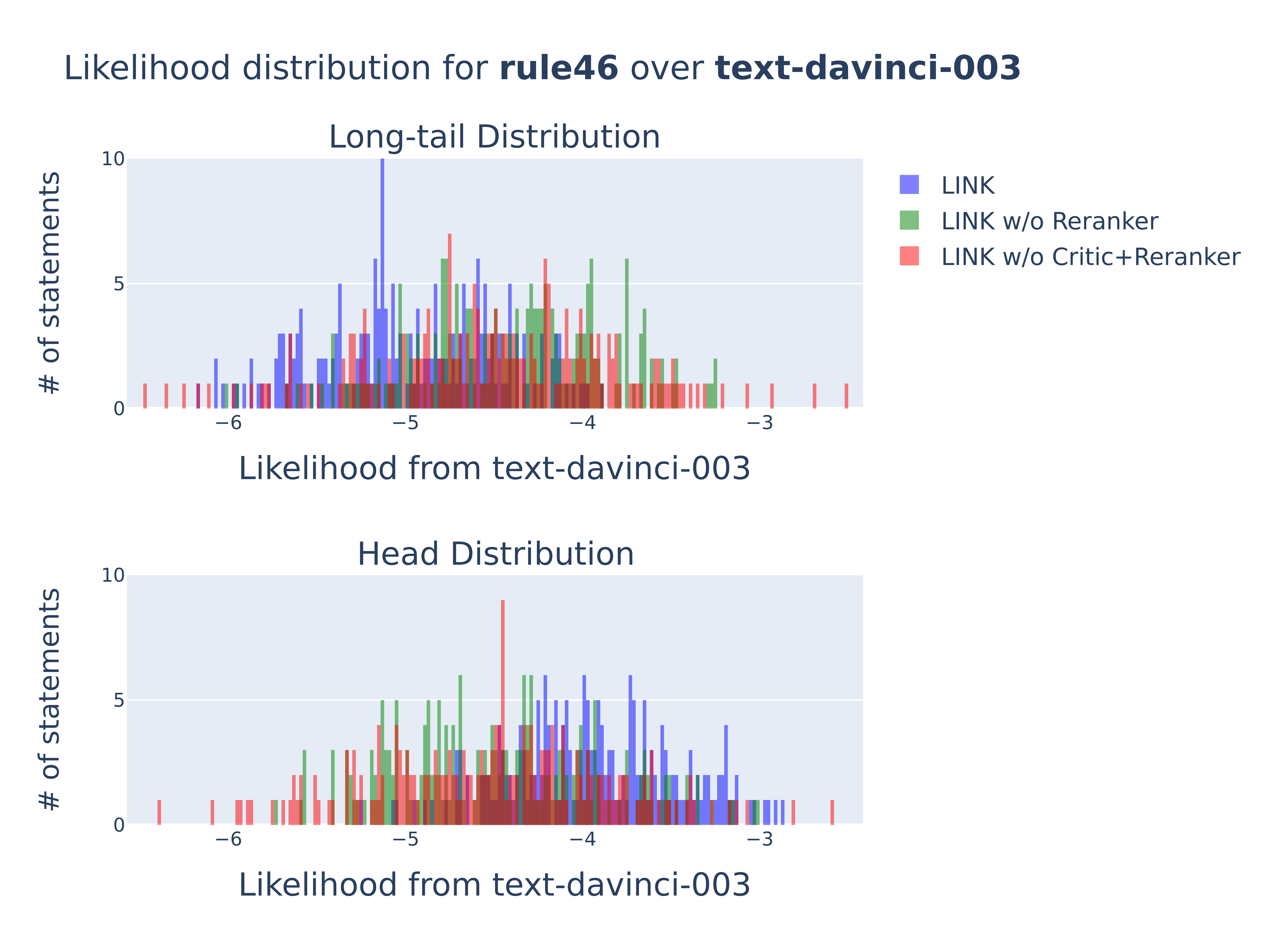}
        \caption{rule46}
    \end{subfigure}
    
    \begin{subfigure}{.3\linewidth}
        \centering
        \includegraphics[width=\textwidth]{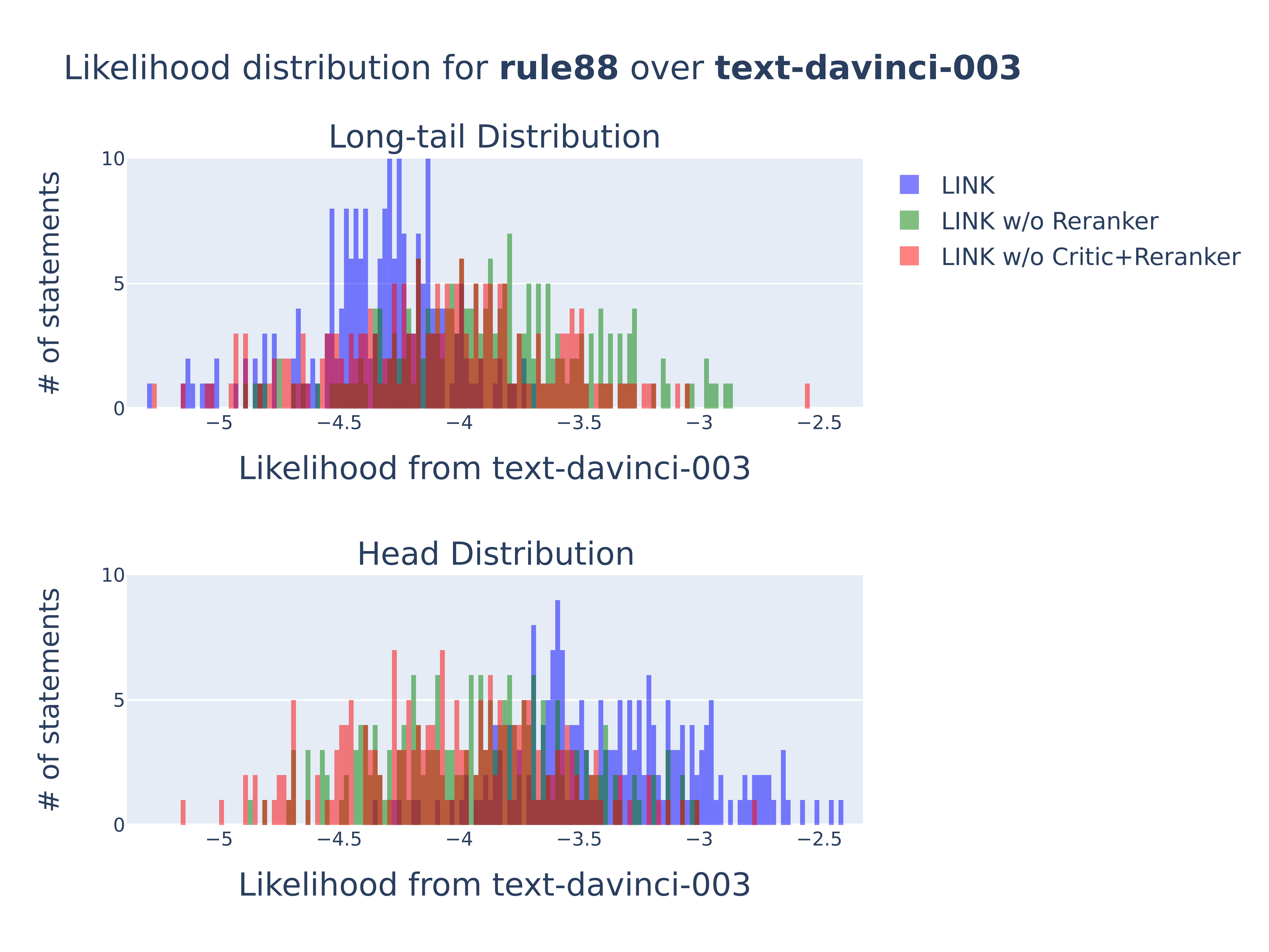}
        \caption{rule88}
    \end{subfigure}
        \hfill
    \begin{subfigure}{.3\linewidth}
        \centering
        \includegraphics[width=\textwidth]{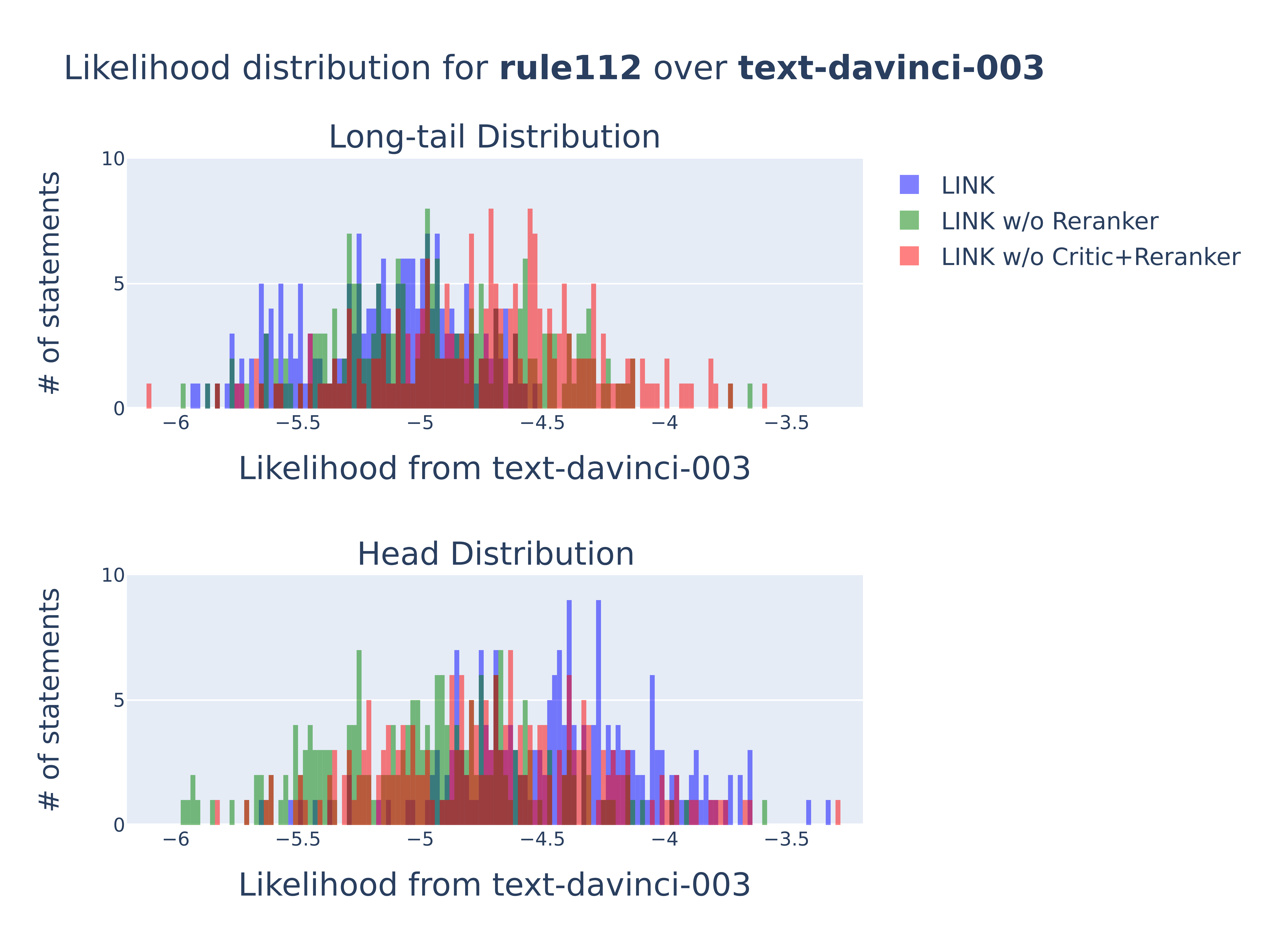}
        \caption{rule112}
    \end{subfigure}
        \hfill
    \begin{subfigure}{.3\linewidth}
        \centering
        \includegraphics[width=\textwidth]{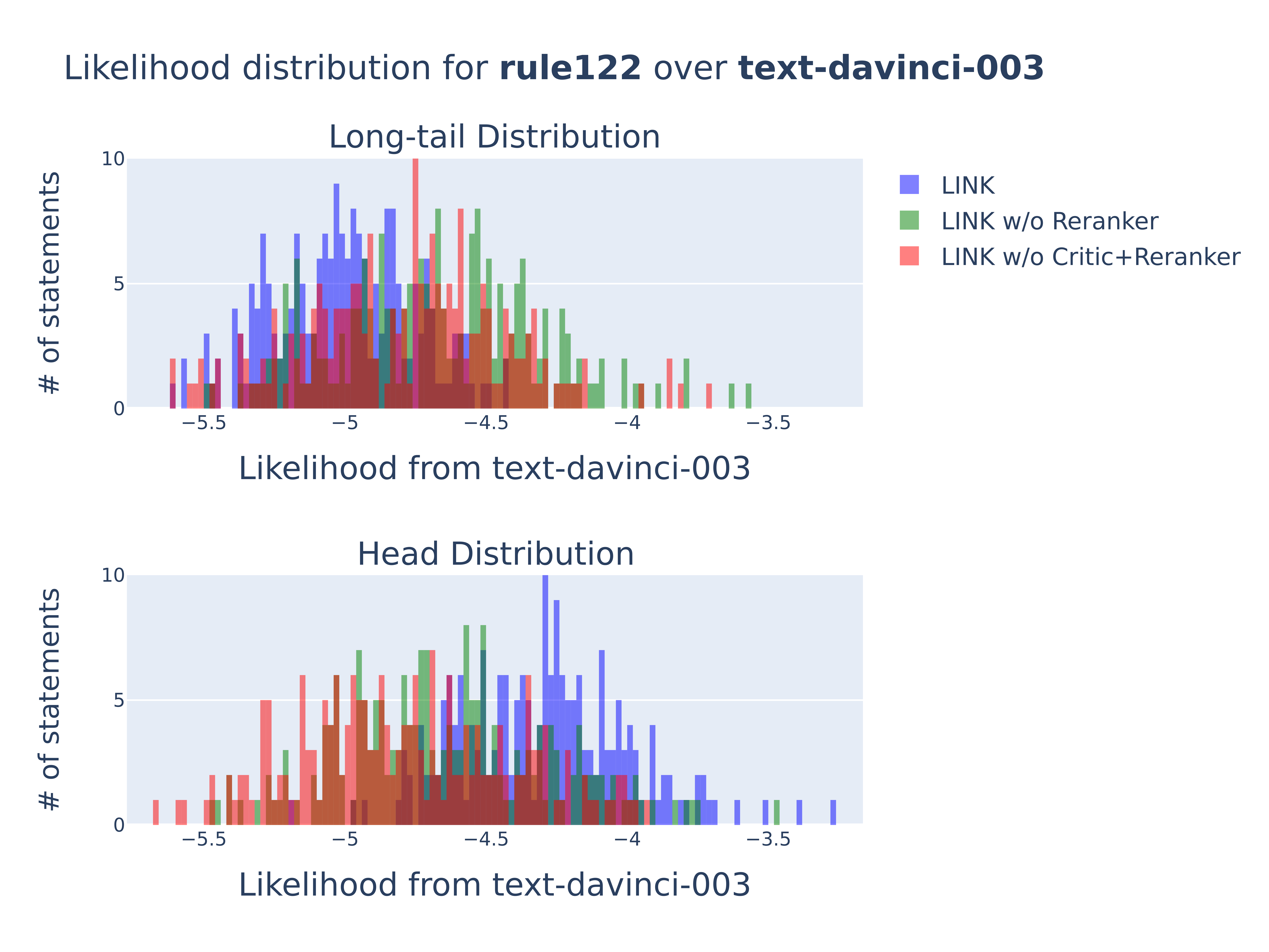}
        \caption{rule122}
    \end{subfigure}
    \caption{When we remove reranker from~\framework, the distribution of the resulting head and \lt statements are pulled towards the center. Using reranker is essential for separating the head and \lt distribution.}
    \label{fig:ablation}
\end{figure*}

To investigate the effectiveness of the reranker, we provide an ablation study on a few sampled rules by replacing the reranker step with a random sampling method.~\Figref{fig:ablation} presents the distribution comparison of generated statements by~\framework~and the variant without the reranker. Without the reranker, the generated statements for both head distribution and \lt distribution are pulled towards the center of the distribution, making them completely inseparable.

\subsection{Ineffectiveness of Post-hoc Reranking for LLM generated knowledge.}

To further highlight the importance of performing step-wise reranking in~\framework, we confirm that applying a post-hoc reranker on the \gpt~generations from instructions does not have the same effect as~\framework. We use \instruct~to rerank the \gpt~generations from the \ti{lowest} to the \ti{highest} likelihood and take the top 75\% results as the \lt distribution. For the head distribution, we rerank the generations from the \ti{highest} to the \ti{lowest} likelihood and take the top 75\% results.

\begin{figure}
    \centering
    \captionsetup{width=\linewidth}
    \includegraphics[width=0.8\linewidth]{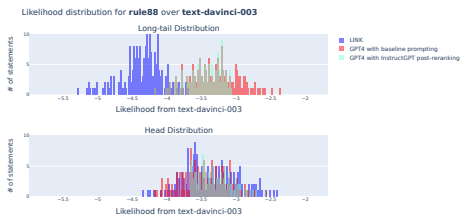}
    \caption{Post-hoc reranking of GPT4 does not help move the distribution towards the \lt distribution.}
    \label{fig:posthoc}
\end{figure}

We evaluate on the same set of rules as in~\Ssecref{subsec:distribution} as an example. \Figref{fig:posthoc} illustrates the distribution of generated statements by \framework, prompt-based \gpt~and prompt-based \gpt~with reranker. We observe that using post-hoc reranker still cannot achieve a separation between the generation of the head distribution and the \lt distribution, even with the same model as the evaluation. It demonstrates that maneuvering the distribution during the searching process is necessary and more effective than post-hoc filtering.

\subsection{Applying \gpt~as the knowledge model}
\label{app:ablation-knowledge}
\begin{figure*}
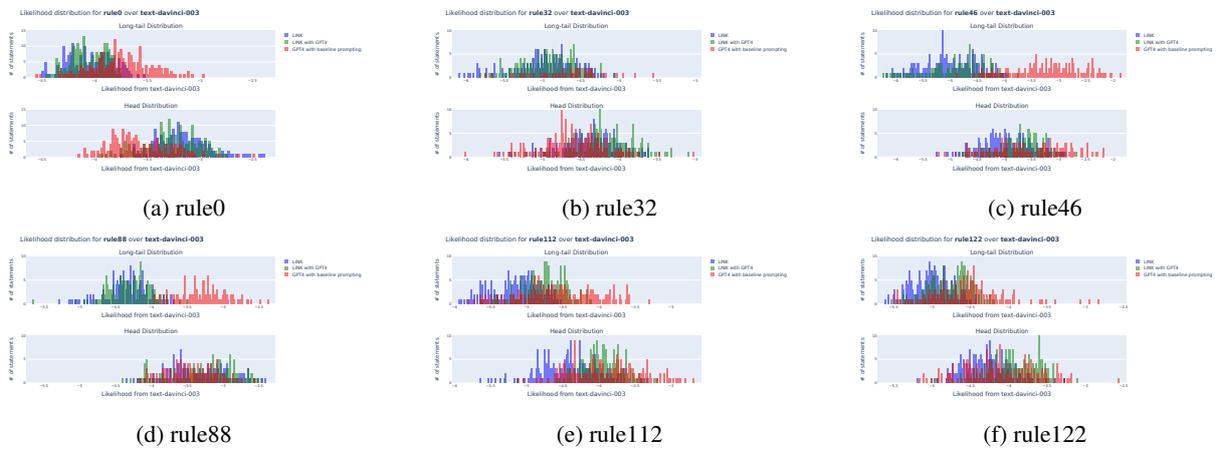

    \centering
    \begin{subfigure}{.3\linewidth}
        \centering
        \includegraphics[width=\textwidth]{images/gpt4/rule0.pdf}
        \caption{rule0}
    \end{subfigure}
        \hfill
    \begin{subfigure}{.3\linewidth}
        \centering
        \includegraphics[width=\textwidth]{images/gpt4/rule32.pdf}
        \caption{rule32}
    \end{subfigure}
        \hfill
    \begin{subfigure}{.3\linewidth}
        \centering
        \includegraphics[width=\textwidth]{images/gpt4/rule46.pdf}
        \caption{rule46}
    \end{subfigure}
    
    \begin{subfigure}{.3\linewidth}
        \centering
        \includegraphics[width=\textwidth]{images/gpt4/rule88.pdf}
        \caption{rule88}
    \end{subfigure}
        \hfill
    \begin{subfigure}{.3\linewidth}
        \centering
        \includegraphics[width=\textwidth]{images/gpt4/rule112.pdf}
        \caption{rule112}
    \end{subfigure}
        \hfill
    \begin{subfigure}{.3\linewidth}
        \centering
        \includegraphics[width=\textwidth]{images/gpt4/rule122.pdf}
        \caption{rule122}
    \end{subfigure}
    \caption{\framework~using GPT4 creates statements that fall in a roughly similar \lt distribution as the original~\framework~with InstructGPT.}
    \label{fig:link_gpt4}
\end{figure*}

\Tabref{tab:link_gpt4} shows the generation quality of \gpt~using baseline prompting method, \framework~and \framework~with \gpt~as the knowledge model over 6 sample rules. Using a stronger model as the knowledge model has marginal effect on the quality of generations compared to \framework. \Figref{fig:link_gpt4} shows that whatever the knowledge model is, the distribution of generations by \framework~can correctly fall in the \lt distribution.

    \begin{table}[ht]
        \centering
        \small
        \caption{Using a stronger model as the knowledge model does not improve generation qualities for \framework, but using \framework~with a language model has significant improvement over zero-shot performance.}
        \begin{tabular}{cccc}
        \toprule
           & Data Type & Factuality & Overall \\
           \midrule
           Zero-shot \gpt & 85.44 & 88.42 & 74.39 \\
           \framework & \textbf{93.42} & \textbf{97.50} & \textbf{91.33} \\
           \framework\ with \gpt & 92.75 & 96.17 & 89.25\\
        \bottomrule
        \end{tabular}
        \label{tab:link_gpt4}
    \end{table}

\section{Addendum on Distribution}
\begin{figure*}
    \centering
    \begin{subfigure}{.3\linewidth}
        \centering
        \includegraphics[width=\textwidth]{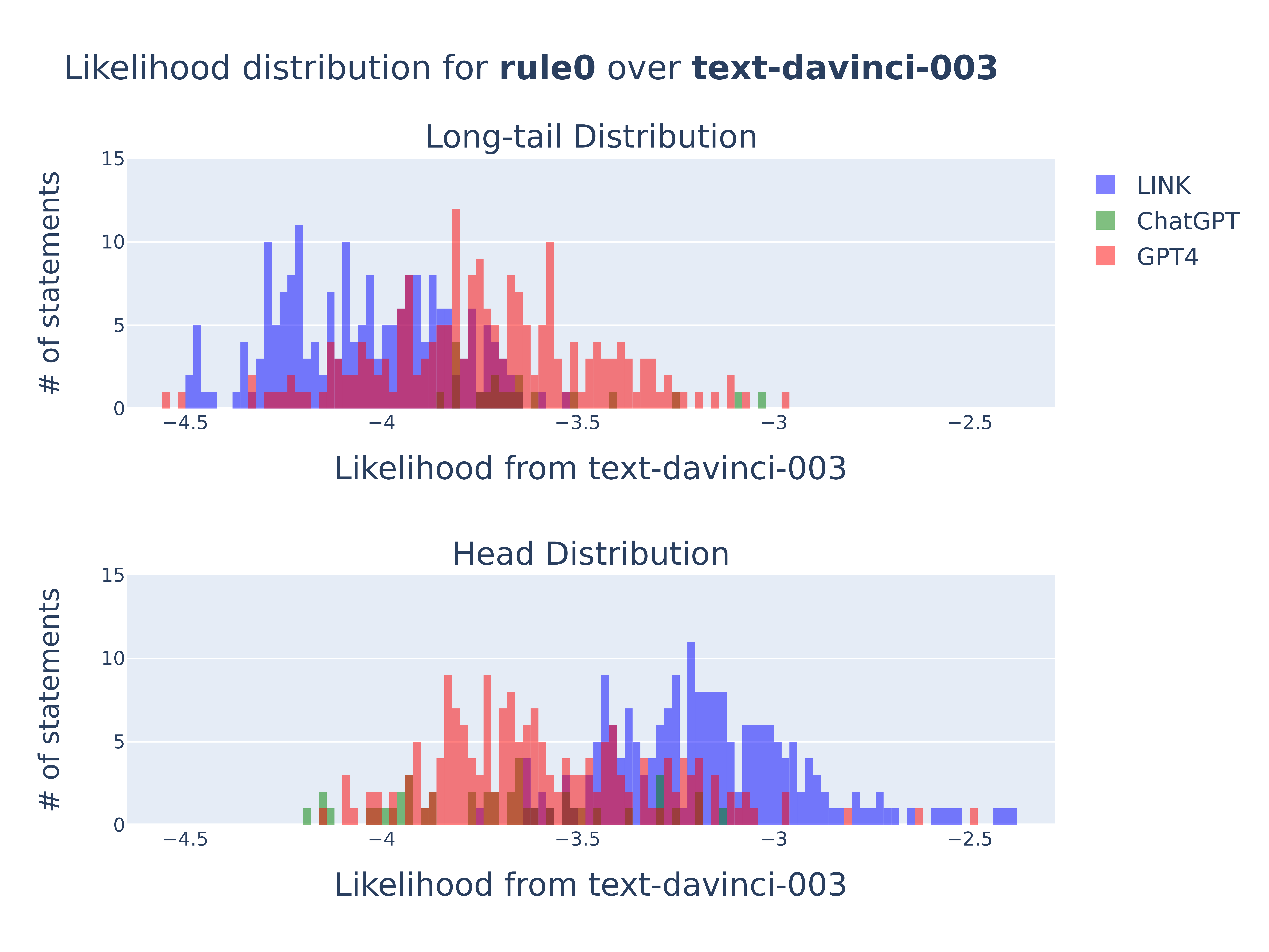}
        \caption{rule0}\label{fig:image11}
    \end{subfigure}
        \hfill
    \begin{subfigure}{.3\linewidth}
        \centering
        \includegraphics[width=\textwidth]{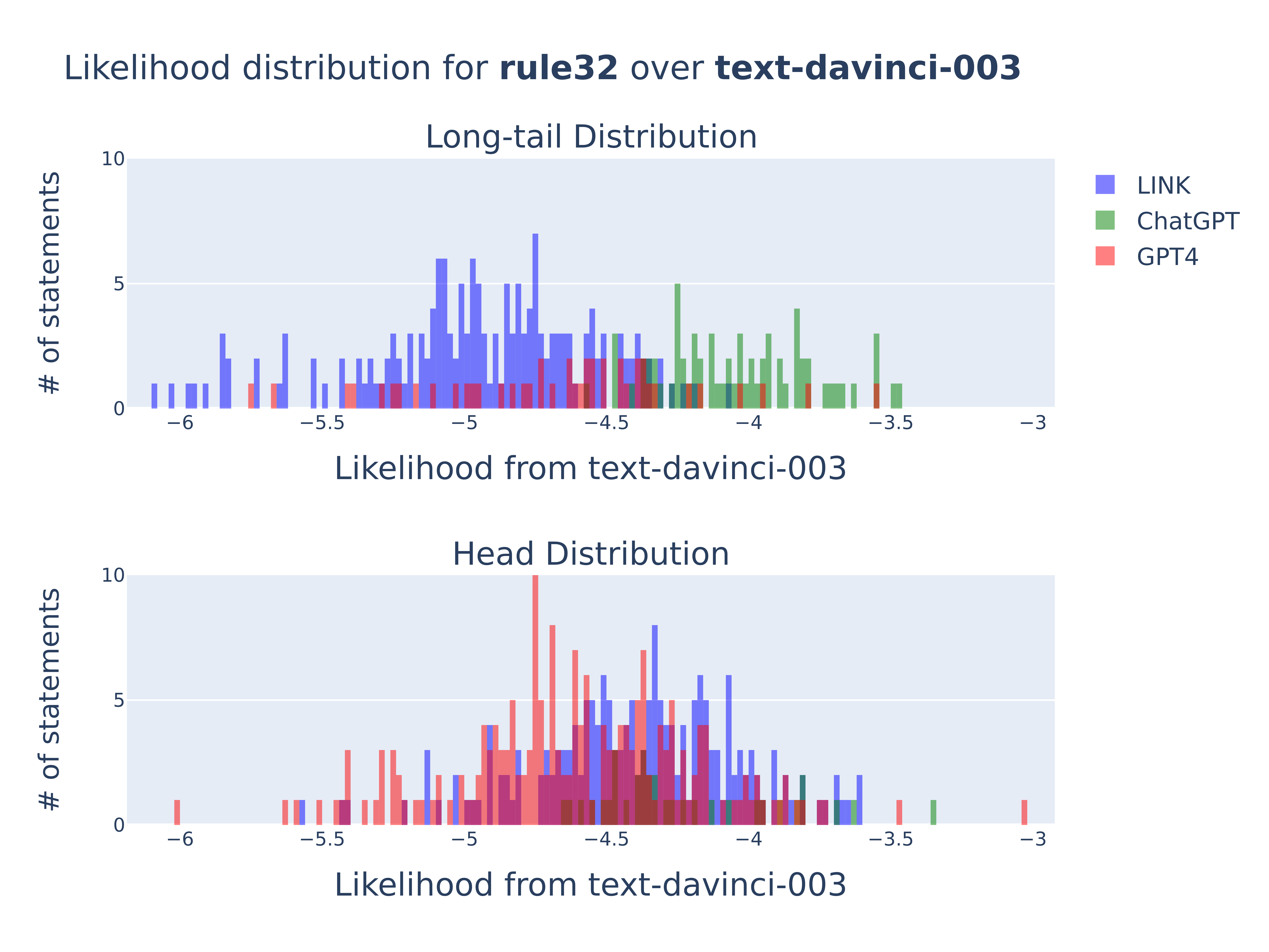}
        \caption{rule32}\label{fig:image12}
    \end{subfigure}
        \hfill
    \begin{subfigure}{.3\linewidth}
        \centering
        \includegraphics[width=\textwidth]{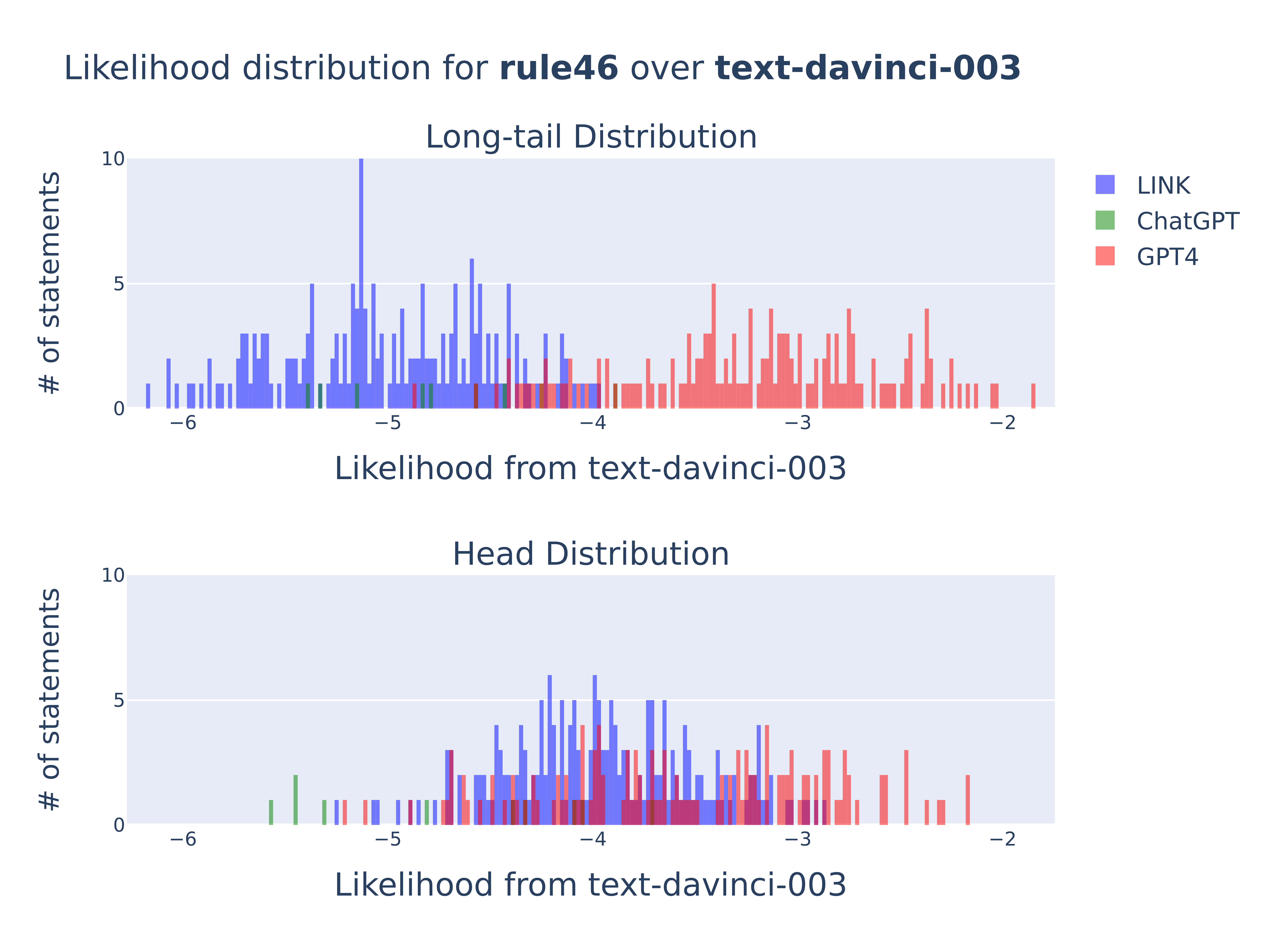}
        \caption{rule46}\label{fig:image13}
    \end{subfigure}
    
    \begin{subfigure}{.3\linewidth}
        \centering
        \includegraphics[width=\textwidth]{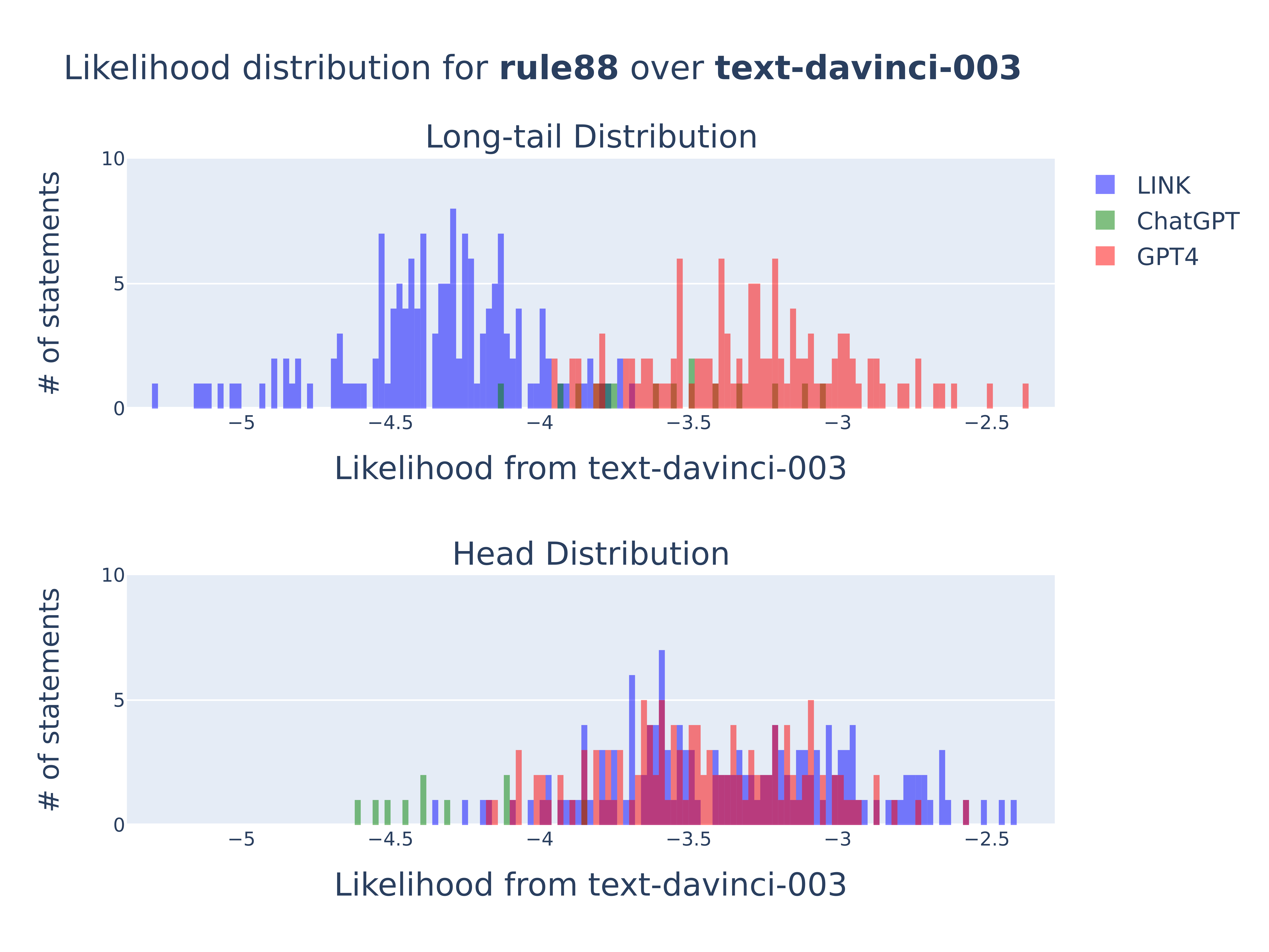}
        \caption{rule88}\label{fig:image21}
    \end{subfigure}
        \hfill
    \begin{subfigure}{.3\linewidth}
        \centering
        \includegraphics[width=\textwidth]{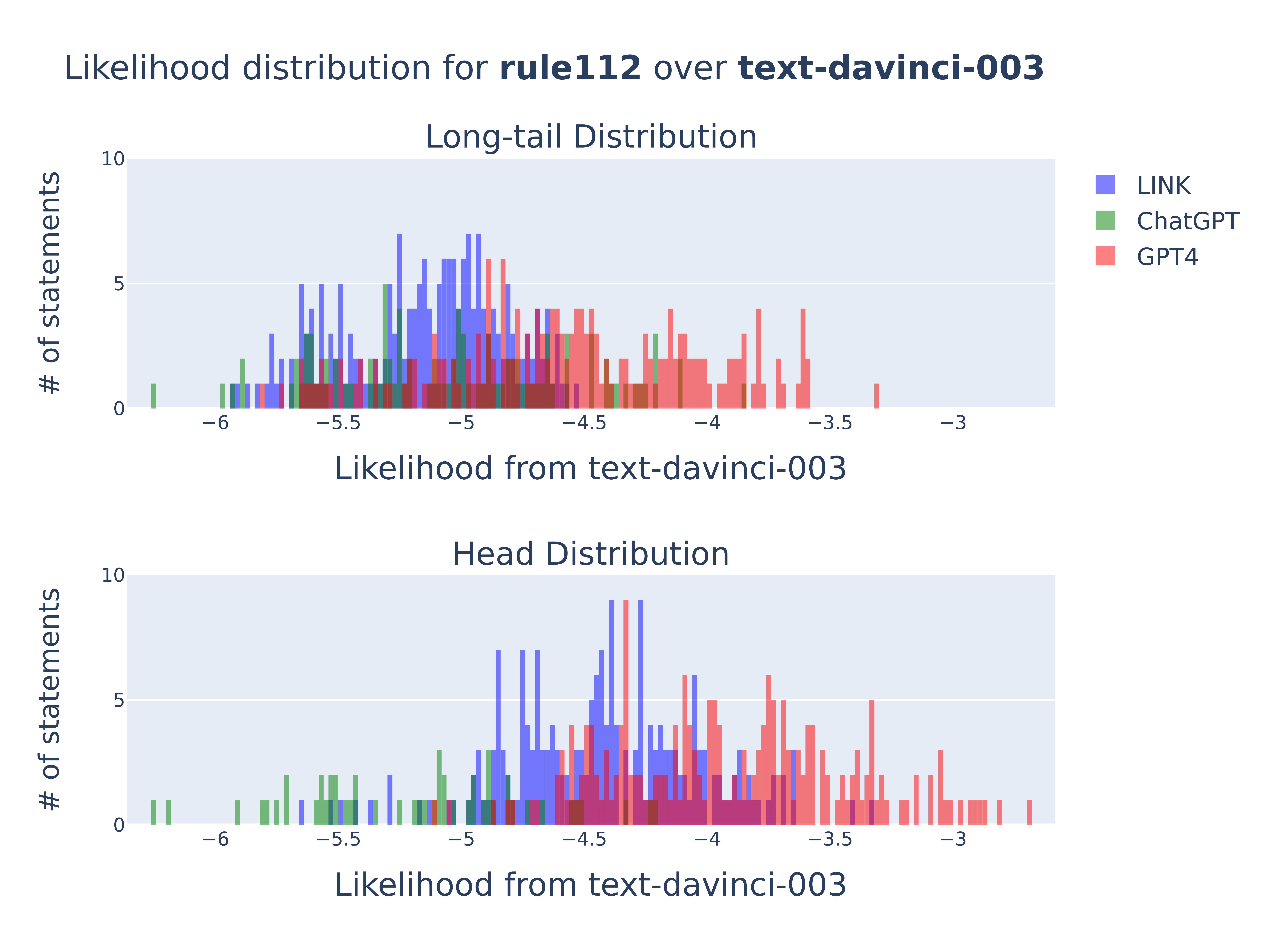}
        \caption{rule112}\label{fig:image22}
    \end{subfigure}
        \hfill
    \begin{subfigure}{.3\linewidth}
        \centering
        \includegraphics[width=\textwidth]{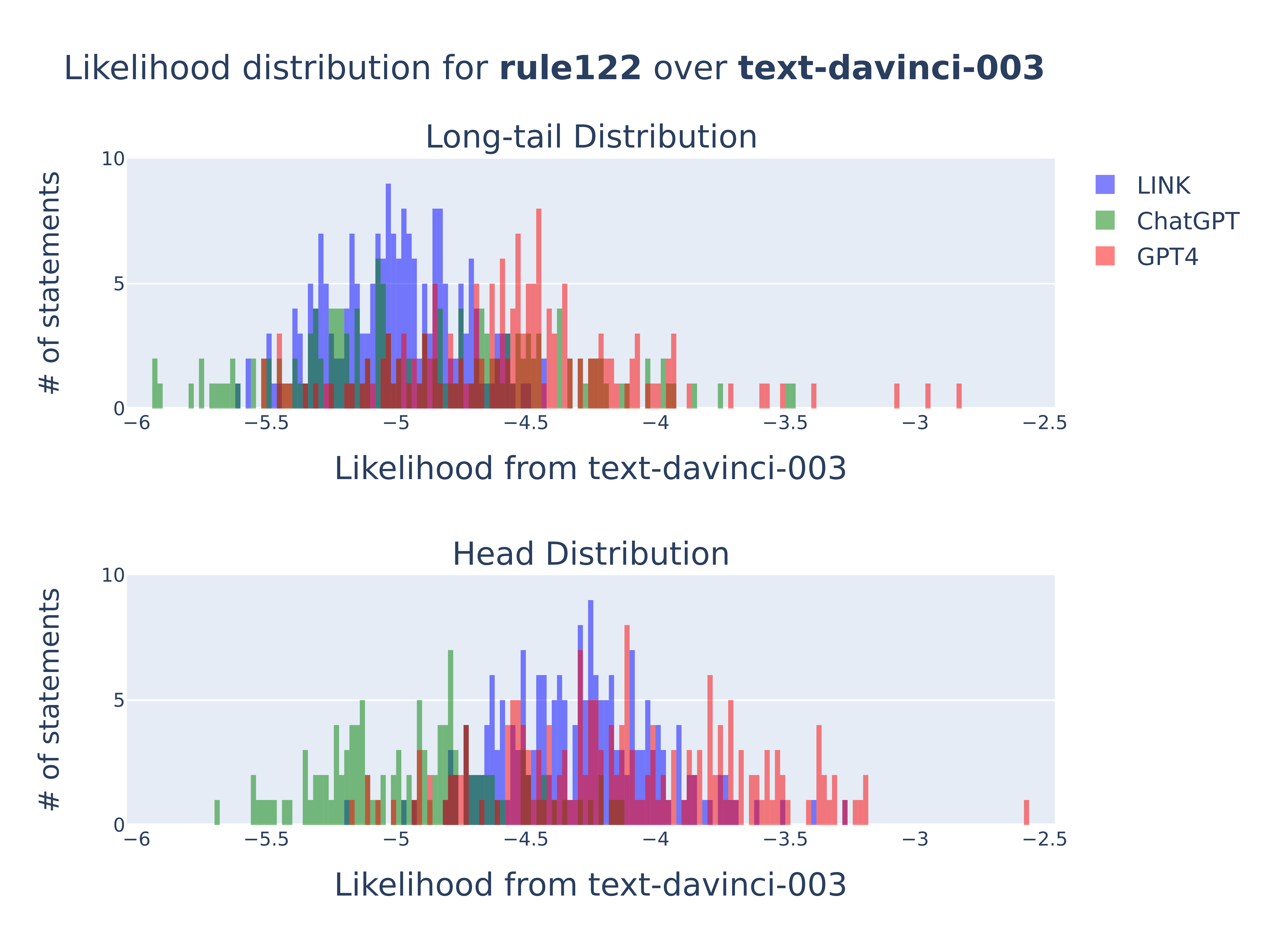}
        \caption{rule122}\label{fig:image23}
    \end{subfigure}
    \caption{An illustration on the distribution of generated statements by \framework, \chat~and \gpt. While \framework's \lt generations fall into a lower probability distribution than those of \gpt, \gpt's ``\lt distribution" overlaps with the head distribution, indicating that these generations are not truly long-tail.}
    \label{fig:multi}
\end{figure*}
\subsection{Additional distribution plots for symbolic rules}
\label{app:more_distribution}
As an extension on \Ssecref{subsec:distribution}, we show the distribution of statements sampled by \framework, \chat~and \gpt~from 6 symbolic rules on \instruct~in \Figref{fig:multi}.

\subsection{Distribution Comparison of Different Models}
\label{app:distribution_comparison}

In this section, we show that the \lt distribution of different language models overlap, and that this evidence supports our assumption that a universal natural language distribution exists; subsequently, the \lt distribution of a language model can be used to approximate the \lt distribution of other language models.

\begin{figure*}
    \centering
    \begin{subfigure}{.3\linewidth}
        \centering
        \includegraphics[width=\textwidth]{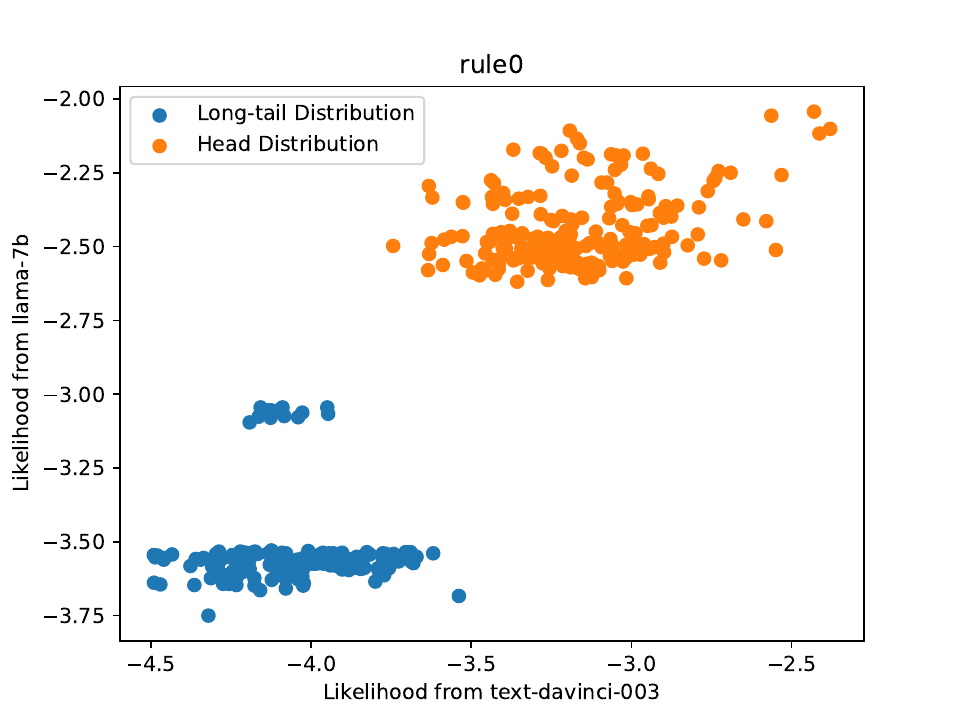}
        \caption{rule0}
    \end{subfigure}
        \hfill
    \begin{subfigure}{.3\linewidth}
        \centering
        \includegraphics[width=\textwidth]{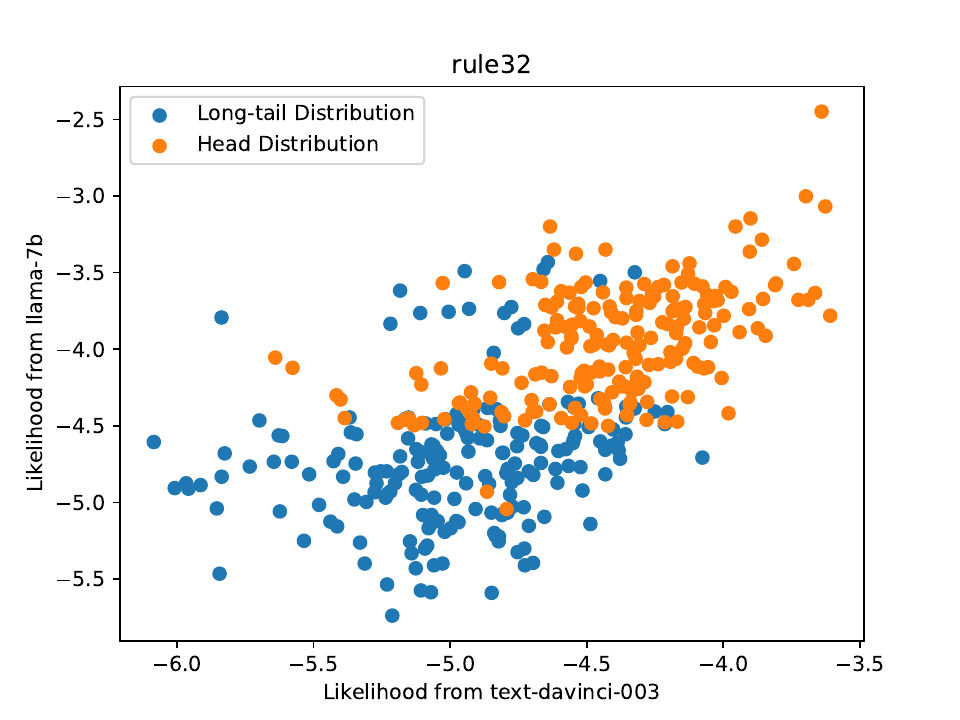}
        \caption{rule32}
    \end{subfigure}
        \hfill
    \begin{subfigure}{.3\linewidth}
        \centering
        \includegraphics[width=\textwidth]{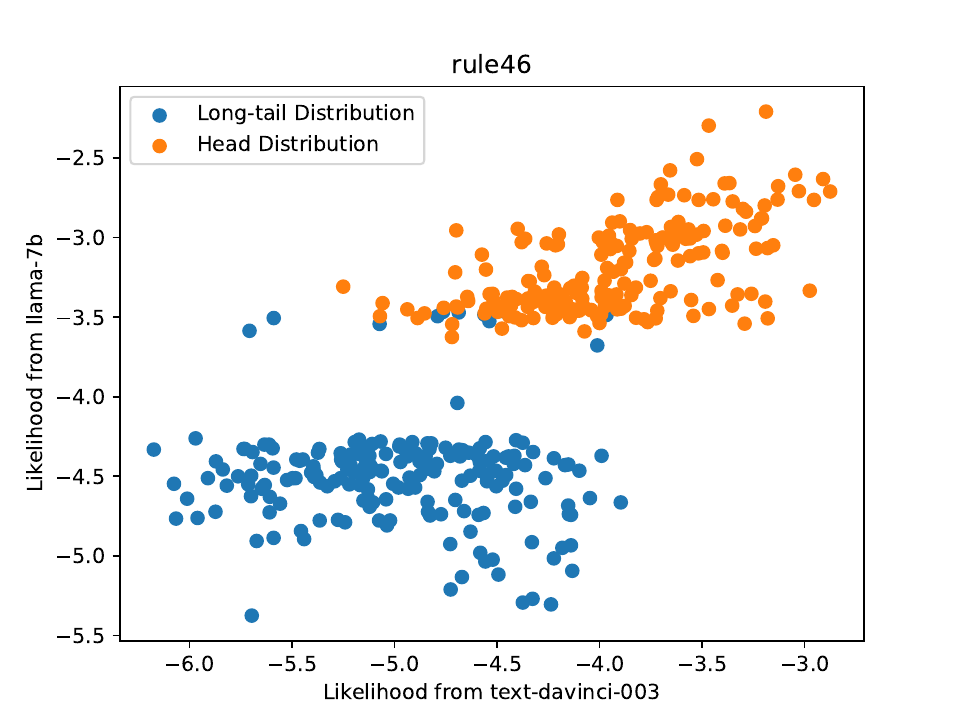}
        \caption{rule46}
    \end{subfigure}
    
    \begin{subfigure}{.3\linewidth}
        \centering
        \includegraphics[width=\textwidth]{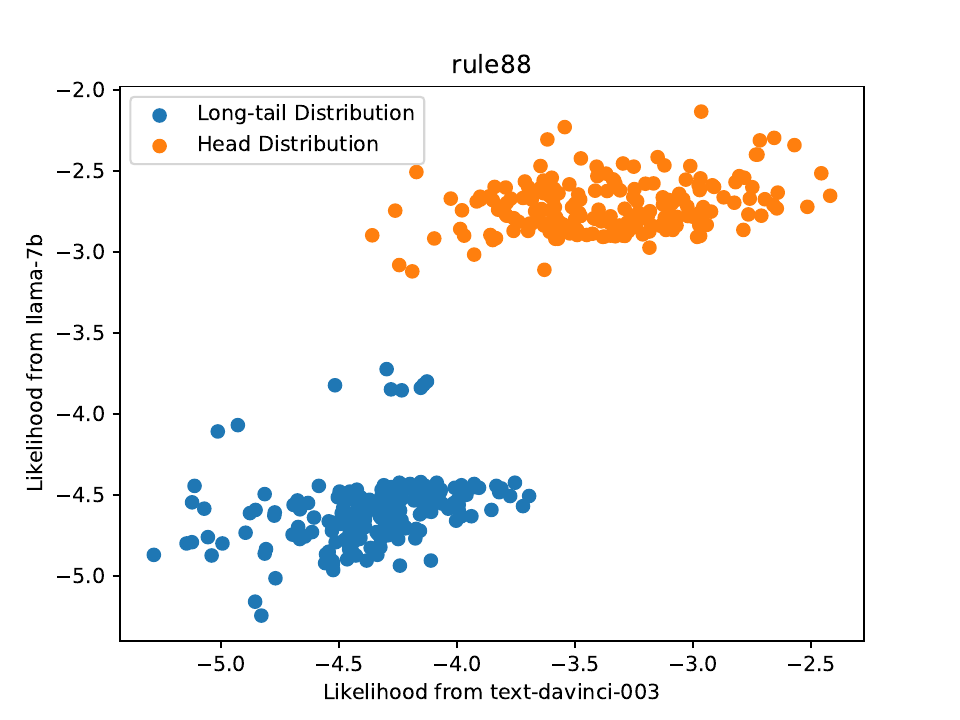}
        \caption{rule88}
    \end{subfigure}
        \hfill
    \begin{subfigure}{.3\linewidth}
        \centering
        \includegraphics[width=\textwidth]{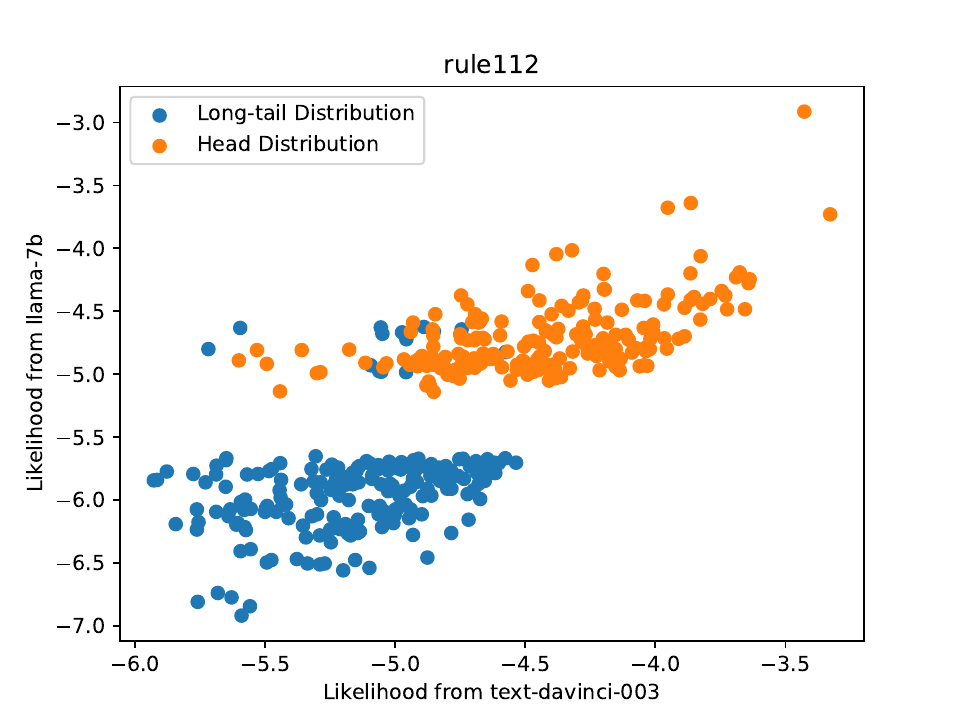}
        \caption{rule112}
    \end{subfigure}
        \hfill
    \begin{subfigure}{.3\linewidth}
        \centering
        \includegraphics[width=\textwidth]{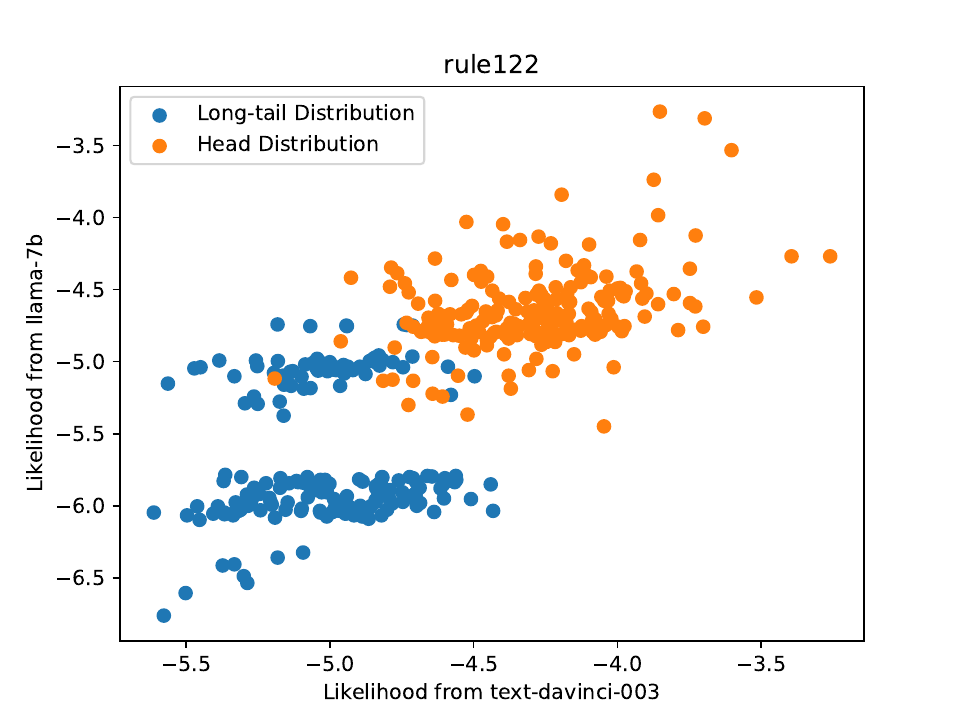}
        \caption{rule122}
    \end{subfigure}
    \caption{An illustration of the distribution comparison between \ttt{llama-7B} and \instruct~of generated statements by \framework.}
    \label{fig:llama_gpt}
\end{figure*}

\begin{figure*}
    \centering
    \begin{subfigure}{.3\linewidth}
        \centering
        \includegraphics[width=\textwidth]{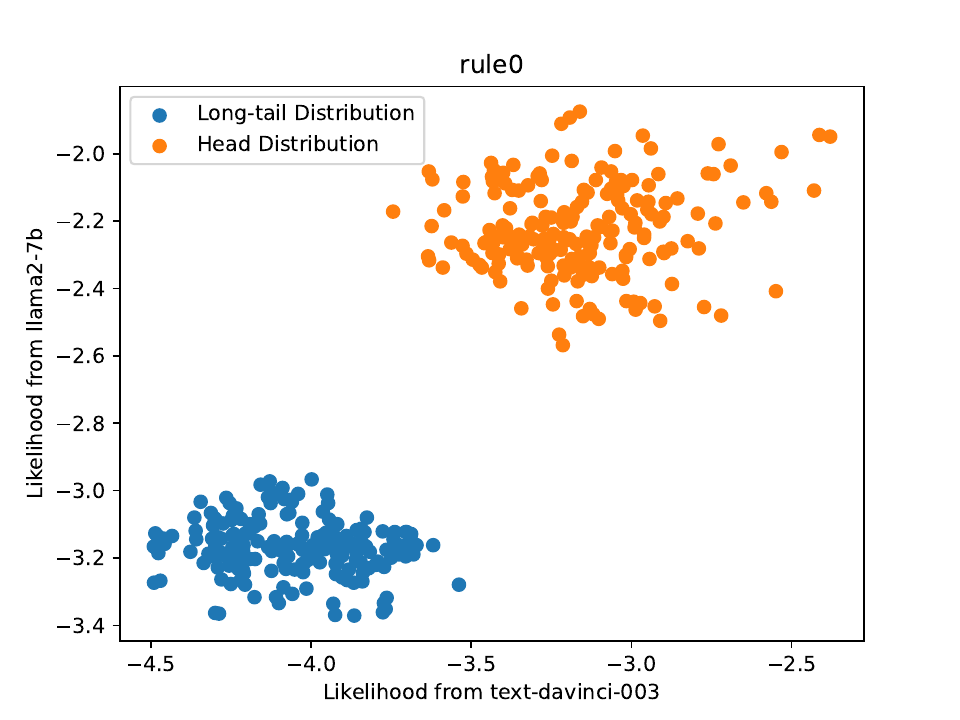}
        \caption{rule0}
    \end{subfigure}
        \hfill
    \begin{subfigure}{.3\linewidth}
        \centering
        \includegraphics[width=\textwidth]{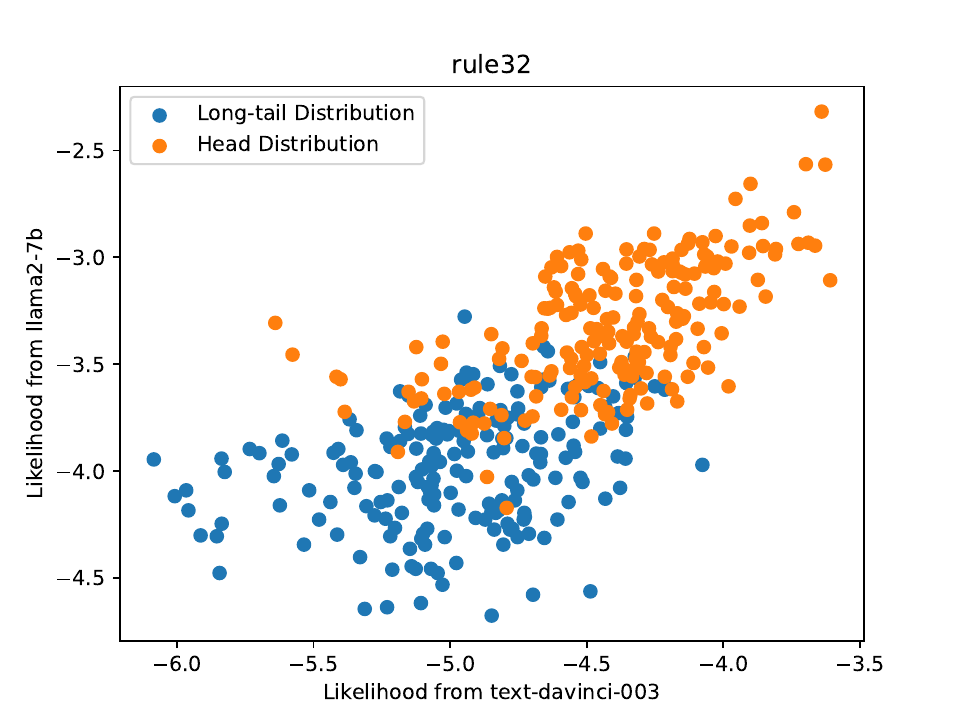}
        \caption{rule32}
    \end{subfigure}
        \hfill
    \begin{subfigure}{.3\linewidth}
        \centering
        \includegraphics[width=\textwidth]{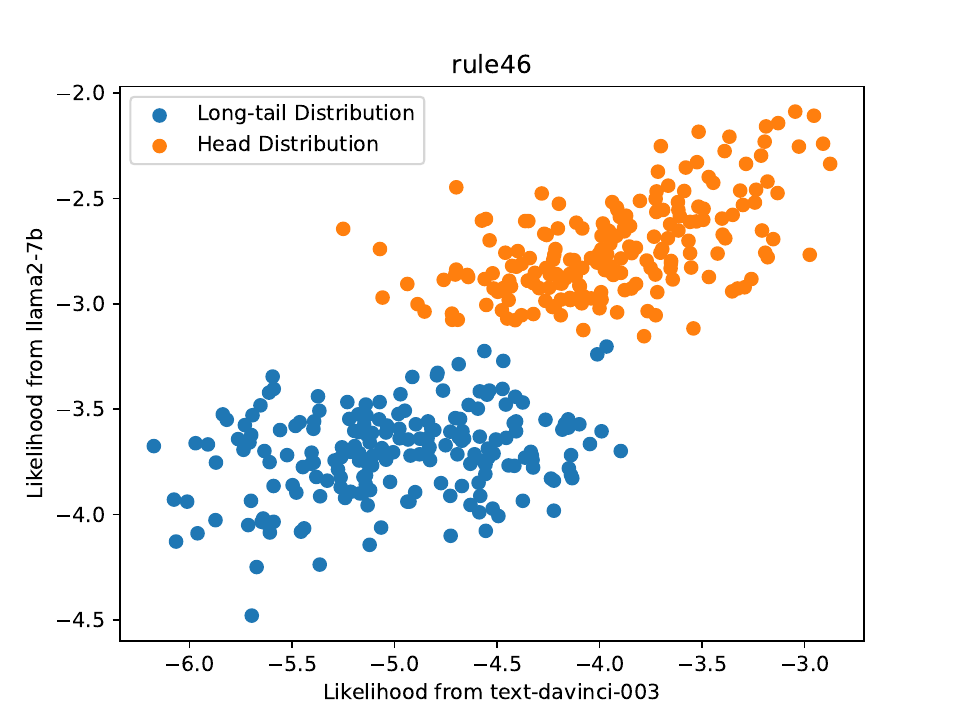}
        \caption{rule46}
    \end{subfigure}
    
    \begin{subfigure}{.3\linewidth}
        \centering
        \includegraphics[width=\textwidth]{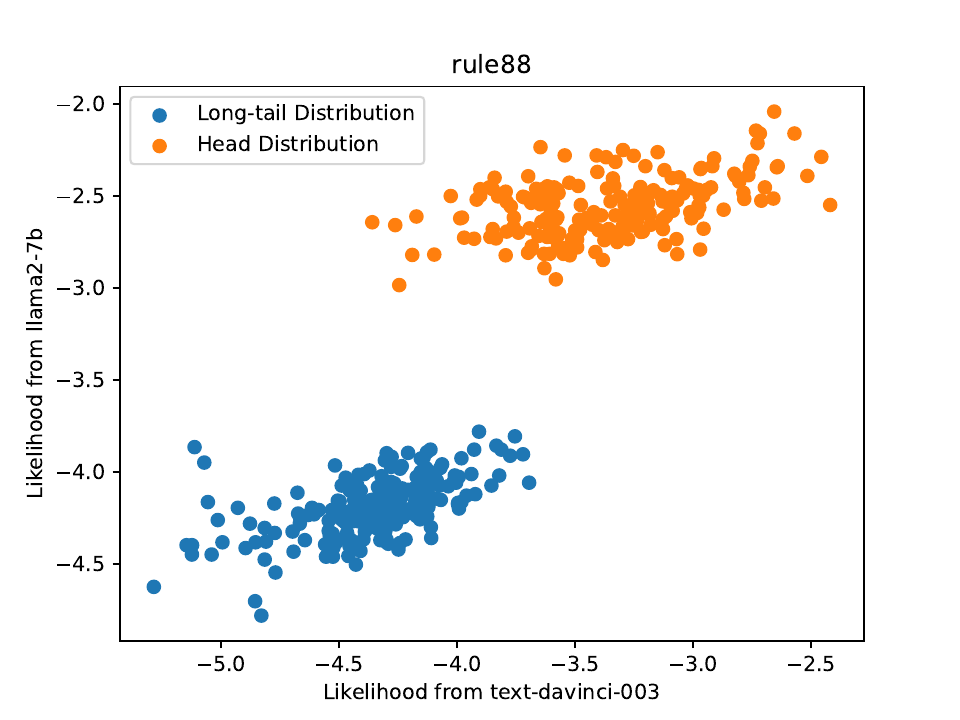}
        \caption{rule88}
    \end{subfigure}
        \hfill
    \begin{subfigure}{.3\linewidth}
        \centering
        \includegraphics[width=\textwidth]{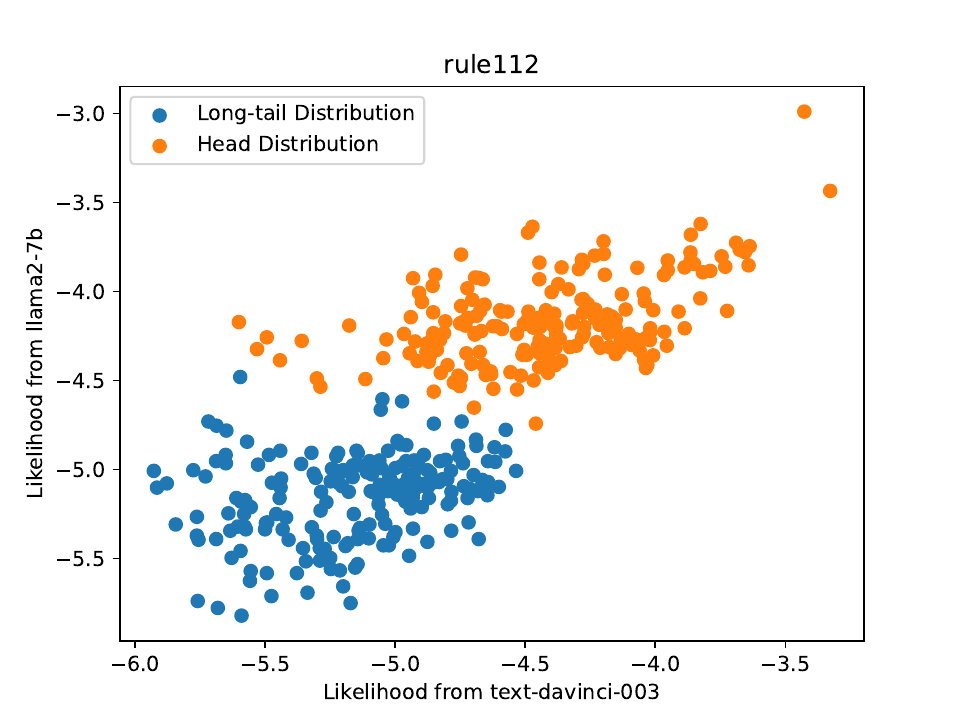}
        \caption{rule112}
    \end{subfigure}
        \hfill
    \begin{subfigure}{.3\linewidth}
        \centering
        \includegraphics[width=\textwidth]{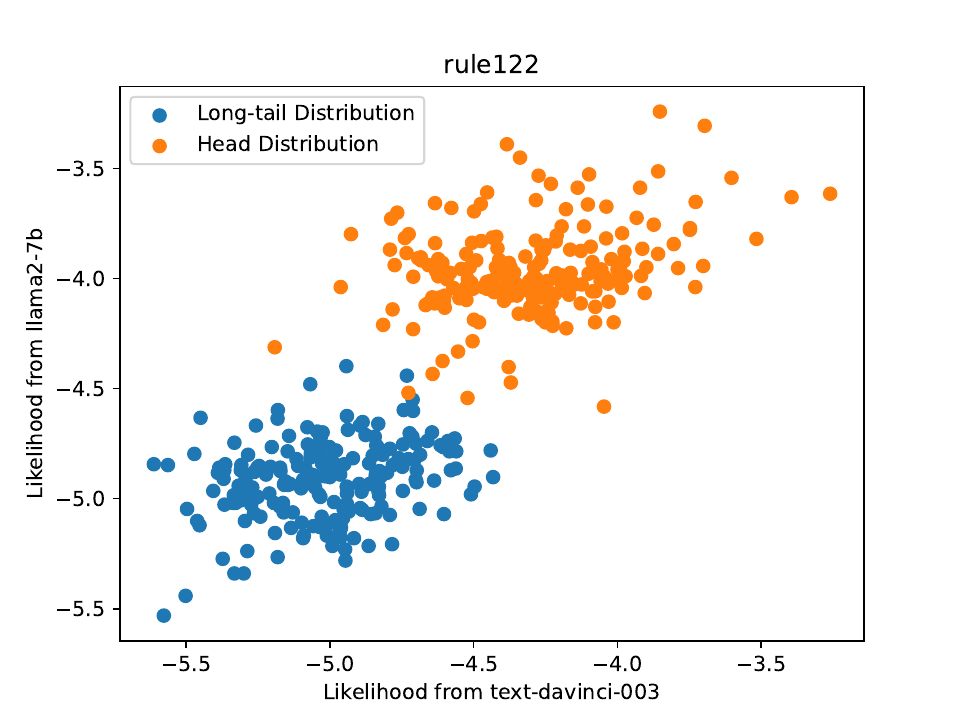}
        \caption{rule122}
    \end{subfigure}
    \caption{An illustration of the distribution comparison between \ttt{llama2-7B} and \instruct~of generated statements by \framework.}
    \label{fig:llama2_gpt}
\end{figure*}

\begin{figure*}
    \centering
    \begin{subfigure}{.3\linewidth}
        \centering
        \includegraphics[width=\textwidth]{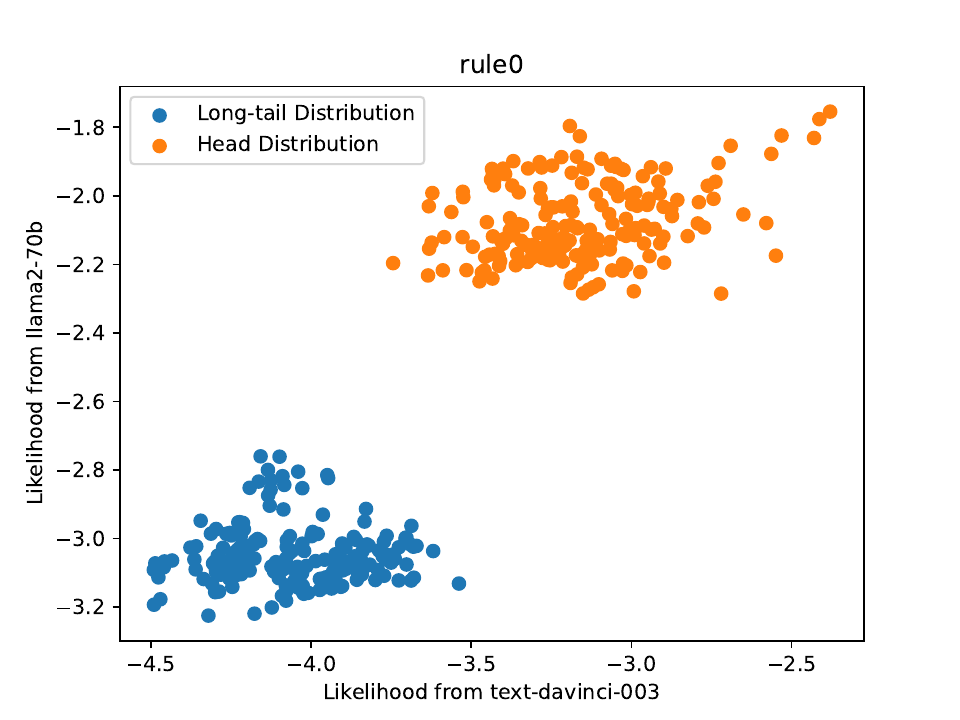}
        \caption{rule0}
    \end{subfigure}
        \hfill
    \begin{subfigure}{.3\linewidth}
        \centering
        \includegraphics[width=\textwidth]{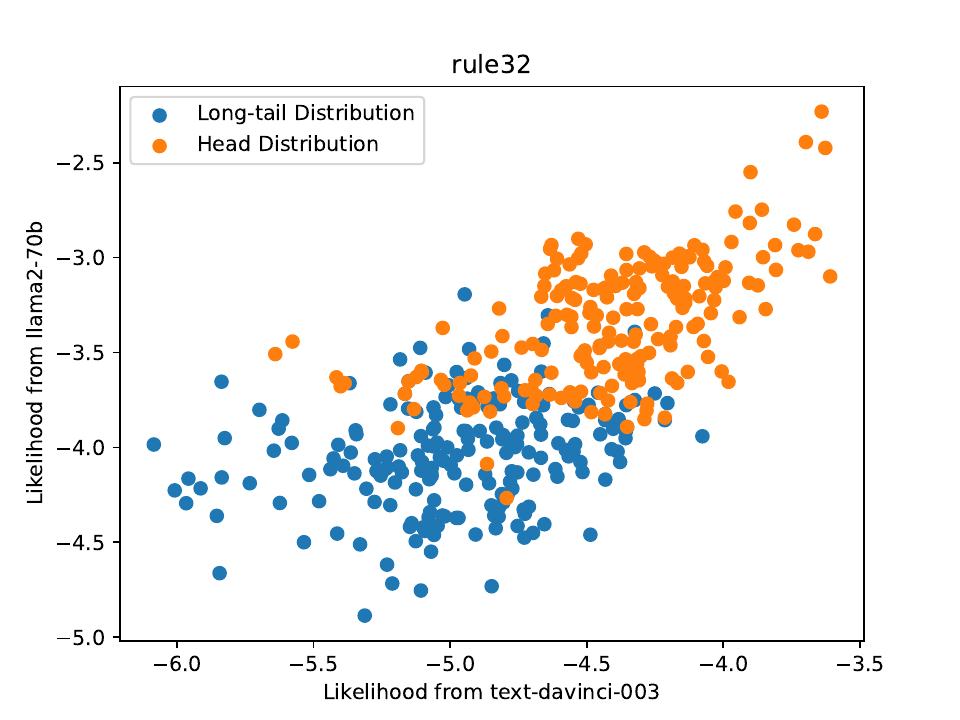}
        \caption{rule32}
    \end{subfigure}
        \hfill
    \begin{subfigure}{.3\linewidth}
        \centering
        \includegraphics[width=\textwidth]{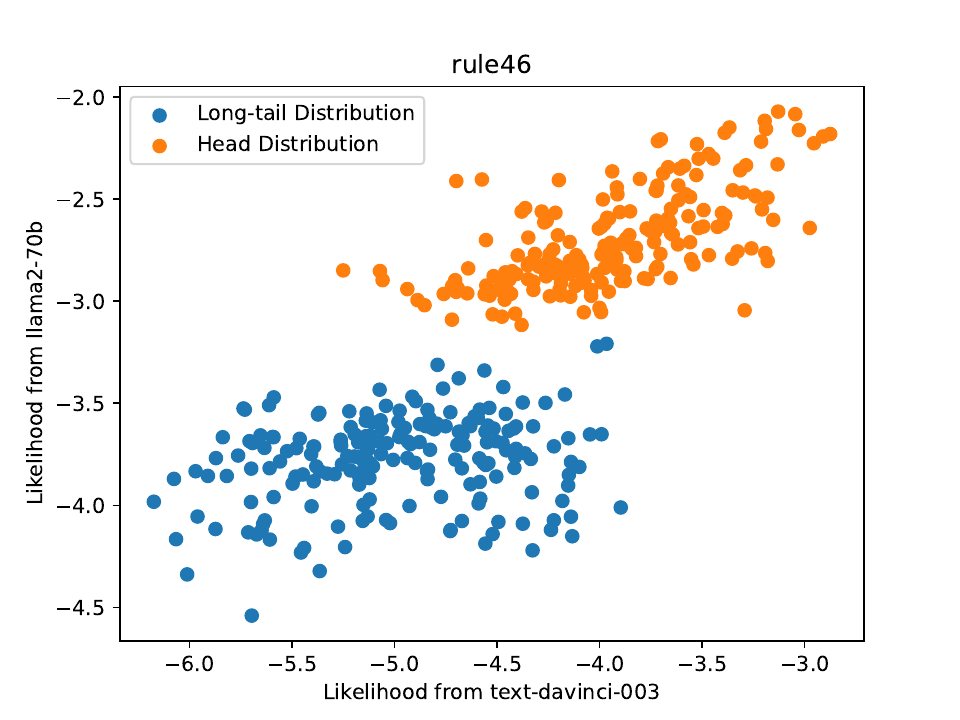}
        \caption{rule46}
    \end{subfigure}
    
    \begin{subfigure}{.3\linewidth}
        \centering
        \includegraphics[width=\textwidth]{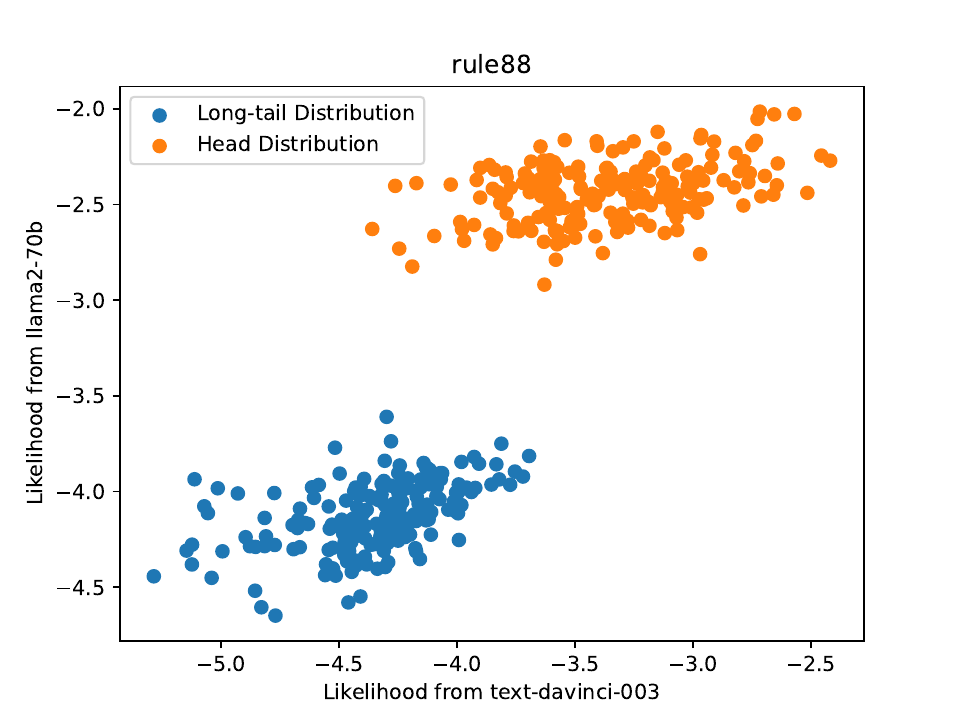}
        \caption{rule88}
    \end{subfigure}
        \hfill
    \begin{subfigure}{.3\linewidth}
        \centering
        \includegraphics[width=\textwidth]{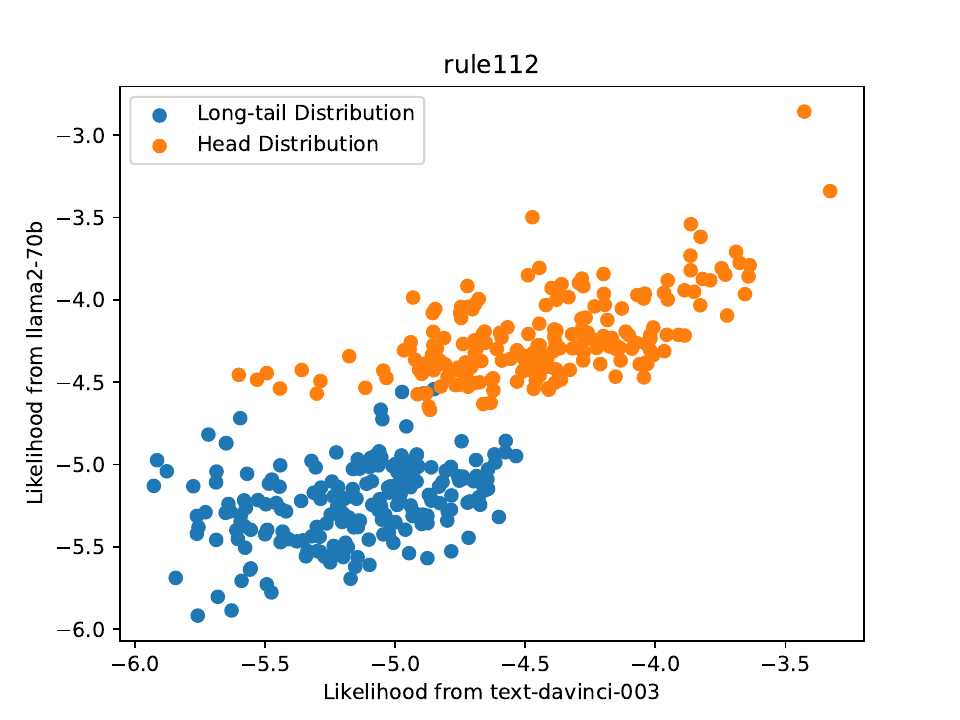}
        \caption{rule112}
    \end{subfigure}
        \hfill
    \begin{subfigure}{.3\linewidth}
        \centering
        \includegraphics[width=\textwidth]{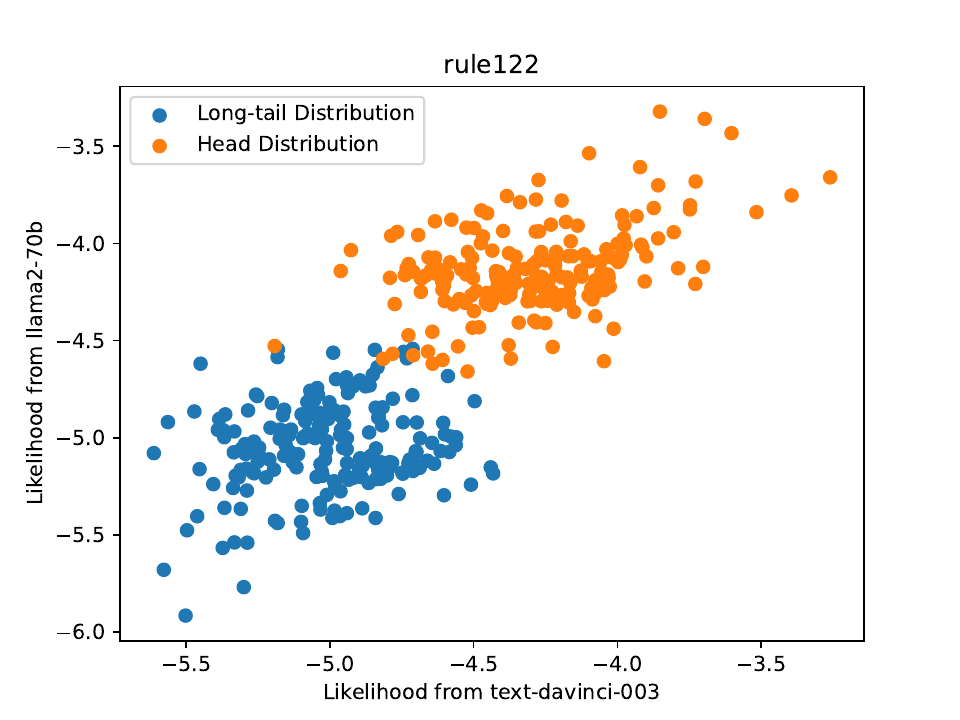}
        \caption{rule122}
    \end{subfigure}
    \caption{An illustration of the distribution comparison between \ttt{llama2-70B} and \instruct~of generated statements by \framework.}
    \label{fig:llama2_70b_gpt}
\end{figure*}

We sample knowledge statements generated by \framework~from six rules and calculate their probabilities with \ttt{llama-7b}, \ttt{llama2-7b}, \ttt{llama-2-70b}, and \instruct. \Figref{fig:llama_gpt}, \Figref{fig:llama2_gpt} and \Figref{fig:llama2_70b_gpt} respectively show the distribution comparison between \instruct~and the three open-source models over the sampled statements from each rule.

For every rule, we note that if a set of statements falls into the low-probability distribution of \instruct, it also falls into the low-probability distribution of the open-source model. Therefore, the categorization on \lt distribution by one language model can effectively approximate the categorization on \lt distribution by other models; hence, we use \instruct~as the approximation of the written natural language distribution in our distribution evaluation.


\section{\Lt Prompts for LLMs}
\label{app:longtail_prompts}
We tried 10 prompts when prompting LLMs to generate knowledge statements in the \lt distribution directly with instructions. Table~\ref{tab:longtail_prompts} shows the 10 prompts which are appended to the original instruction.

\begin{table}[h]
    \centering
    \small
    \caption{An illustration of 10 prompts that instruct LLMs to generate knowledge statements in the \lt distribution.}
    \begin{tabular}{p{.1\textwidth}p{.7\textwidth}}
        \toprule
         & Prompt \\
         \midrule
        1 & Use less frequent terms of A and B and Z.\\
        2 & Use terms of A and B and Z that are less common.\\
        3 & Use terms with lower frequency for A and B and Z.\\
        4 & Use terms of A and B and Z that have lower probability in language model distribution.\\
        5 & Use less frequent words of A and B and Z.\\
        6 & Use words of A and B and Z that are less common. \\
        7 & Use words with lower frequency for A and B and Z. \\
        8 & Use less frequent entities of A and B and Z. \\
        9 & Use entities of A and B and Z that are less common. \\
        10 & Use entities with lower frequency for A and B and Z. \\
        \bottomrule
    \end{tabular}
    \label{tab:longtail_prompts}
\end{table}

\begin{figure}
    \centering
    \captionsetup{width=\linewidth}
    \includegraphics[width=0.9\linewidth]{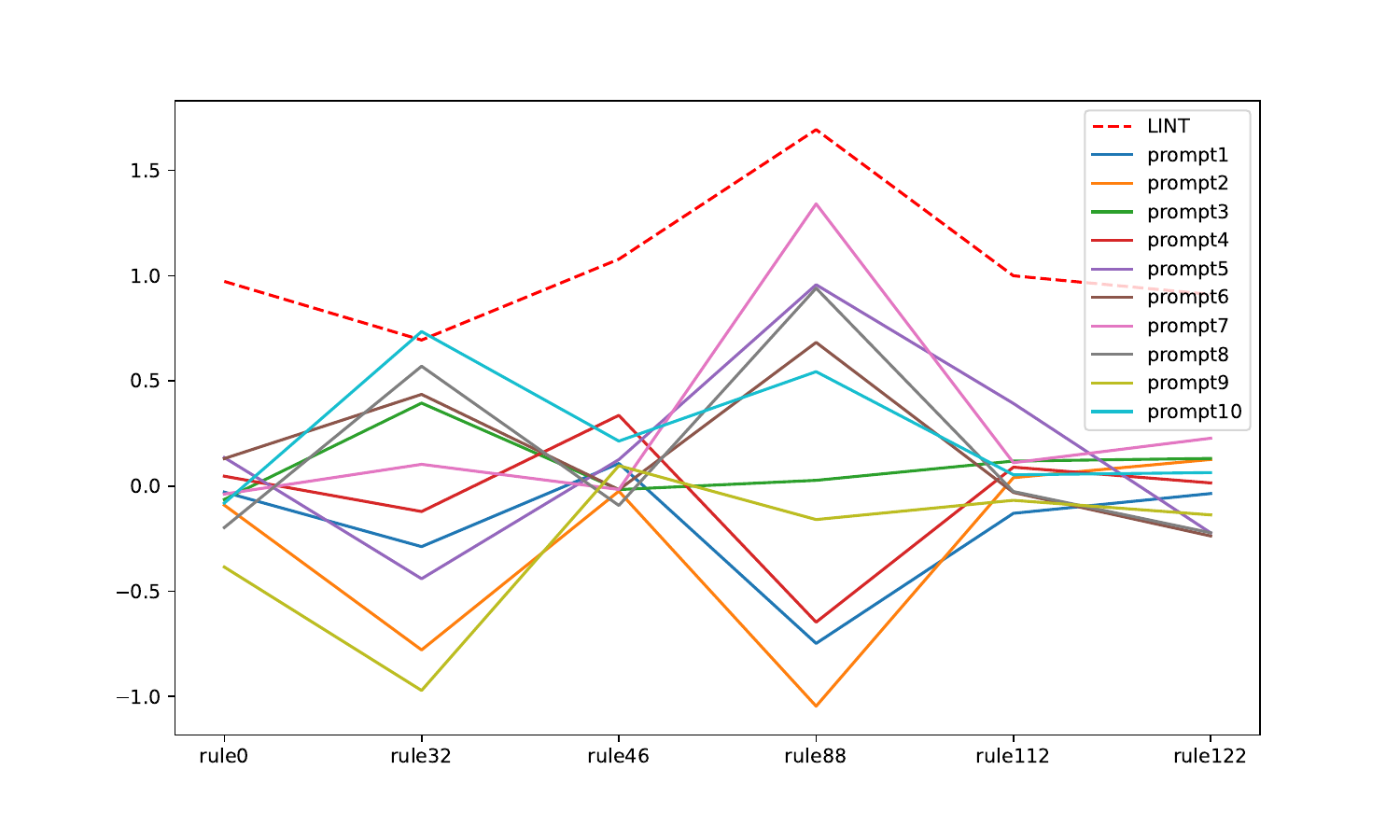}
    \caption{\dataset~has a higher $\delta$ than \gpt~baselines with 10 different prompts.}
    \label{fig:longtail_prompts}
\end{figure}

Following Section~\ref{subsec:distribution}, Figure~\ref{fig:longtail_prompts} shows the $\delta$ value for \gpt~baselines with these prompts and \dataset. These prompts have a similar effect on the distribution of generated statements on most rules and \dataset~consistently has a higher $\delta$, indicating that no matter the prompt we use, LLMs cannot directly generate \lt statements by following instructions. Though these prompts have different $\delta$ values on some rules, no one can consistently achieve a higher $\delta$ among them.


\section{Entailment Classification Probing}
\subsection{Probing template}
\label{app:probing_template}

\Tabref{tab:probing_templates} shows templates we used for the entailment classification task. As mentioned in \Ssecref{sec:entailment}, we divide the templates into positive templates and negative templates. Positive templates are those with a positive label (i.e., \ti{Yes}, \ti{Right} and \ti{True}) and negative templates are those with a negative label (i.e., \ti{No}, \ti{Wrong} and \ti{False}).

Most of the templates have definite labels across all rules. However, the label of Template 7 depends on the rules. If the rule has a positive conclusion (e.g., \ti{Person X can use ChatGPT}), the answer to the question should be positive, i.e., \ti{Yes}. On the contrary, if the rule has a negative conclusion(e.g., \ti{Person X cannot use ChatGPT}), the answer to the question should be negative, i.e., \ti{No}.

\begin{table}[ht]
    \centering
    \small
    \caption{Templates used for machine entailment classification task.
    }
    \begin{tabular}{clcc}
    \toprule
       & Template & Label \\
     \midrule
    1 & \makecell[l]{Is it true that if \ttt{premise}, \ttt{conclusion}.} & Yes\\
    2 & Yes or no: if \ttt{premise}, \ttt{conclusion}. & Yes\\
    3 & True or false: if \ttt{premise}, \ttt{conclusion}. & True\\
    4 & \makecell[tl]{Right or Wrong: if \ttt{premise}, \\ \ttt{conclusion}.} & Right\\
    5 & \makecell[tl]{Premise: \ttt{premise}. \\Conclusion: \ttt{conclusion}. \\Does premise entail conclusion?} & Yes\\
    6 & \makecell[tl]{Premise: \ttt{premise}. \\ Conclusion: \ttt{conclusion\_negation}. \\Does premise contradict the conclusion?} & Yes\\
    7 & \makecell[tl]{Answer the question with yes or no: \\if \ttt{premise}, \ttt{conclusion\_question}?} & \ti{Depends}\\
    8 & \makecell[tl]{Is it true that if \ttt{premise}, \\ \ttt{conclusion\_negation}.} & No\\
    9 & \makecell[tl]{Yes or no: if \ttt{premise}, \\ \ttt{conclusion\_negation}.} & No\\
    10 & \makecell[tl]{True or false: if \ttt{premise}, \\ \ttt{conclusion\_negation}.} & False\\
    11 & \makecell[tl]{Right or Wrong: if \ttt{premise}, \\ \ttt{conclusion\_negation}.} & Wrong \\
    12 & \makecell[tl]{Premise: \ttt{premise}. \\ Conclusion: \ttt{conclusion\_negation}.\\ Does premise entail conclusion?} & No \\
    13 & \makecell[tl]{Premise: \ttt{premise}. \\ Conclusion: \ttt{conclusion}. \\Does premise contradict the conclusion?} & No\\
    \bottomrule
    \end{tabular}
    \label{tab:probing_templates}
\end{table}





\subsection{Accuracy averaged over templates}
One main concern of not using accuracy of our main metric is because accuracy is not favorable for label imbalance. Because we have 13 templates where for each knowledge statement we will have an uneven number of positive or negative answers, the negative label rate in total is around 51.58\%. Therefore, we decide to consider the template biases and deem a model correct only if it answers all templates of a statement correctly, as our goal is to test the true knowledge of models. 

When evaluating the accuracy across all templates, the model's performance also drops over \lt knowledge. \Tabref{tab:eval_gpt4} shows the accuracy across all templates of \gpt. Even if LLMs are brittle to templates, they exhibit a performance drop in the \lt distribution among all domains.

\begin{table}[ht]
    \centering
\begin{tabular}{lcccc} 
    \toprule 
    & Head & Long-tail \\ 
    \midrule
    Natural Properties & 67.07 & 63.40 \\ 
    Temporal & 83.14 & 78.97 \\ 
    Outcomes and Effects & 81.79 & 80.19 \\ 
    Locational & 68.90 & 67.07 \\ 
    \bottomrule 
\end{tabular}
    \caption{Accuracy over all templates, \gpt}
    \label{tab:eval_gpt4}
\end{table}

\subsection{Model and human error analysis}
\Tabref{tab:error_analysis} shows probing examples in which models and humans make error. For those from the head distribution, the entailment can be easy if humans can use search engines. For example, humans can search the periods of “The Paleolithic Era” and “Lion Gate of Mycenae”, and answer easily. Thus the human errors in the head distribution may be due to carelessness. For those containing \lt knowledge, even with search engines, it is not so easy to infer the answer for human annotators. It is also likely that models do not have such knowledge either.

\begin{table*}[h]
    \centering
    \small
    \caption{Examples that human and model make mistakes.}
    \begin{tabular}{p{.3\textwidth}p{.6\textwidth}}
        \toprule
        \textbf{Statement:} & Yes or no: if Plant X vanished in Paleolithic Era, Plant X cannot surround Lion Gate of Mycenae. (Head)\\
        \textbf{True label:} & Yes \\
        \textbf{Human answer:} & No \\
        \textbf{Model answer (GPT4 + COT):} & Yes. The Paleolithic Era ended around 10,000 BC and the Lion Gate of Mycenae was built around 1250 BC. If PlantX vanished in the Paleolithic Era, it could not have surrounded the Lion Gate of Mycenae which was built much later. \\
        \midrule
        \textbf{Statement:} & Yes or no: if Bag X has trouble containing Pandeiro, Dhak cannot fit in Bag X. (\Lt)\\
        \textbf{True label:} & Yes \\ 
        \textbf{Human answer:} & No \\
       \textbf{Model answer (GPT4 + COT):} & No. The size of the Pandeiro does not directly relate to the size of the Dhak. They are different objects and may have different shapes and sizes. Therefore, just because BagX has trouble containing a Pandeiro doesn't necessarily mean that a Dhak cannot fit in BagX. \\
    \bottomrule
    \end{tabular}
    \label{tab:error_analysis}
\end{table*}

\subsection{Rationale analysis}

As mentioned in \Ssecref{sec:entailment}, we examine the rationale the model generated during COT in the entailment classification task and found that the models tend to avoid drawing a ``definite conclusion''. \Tabref{tab:rationale} shows an example.

\begin{table}[ht]
    \centering
    \small
    \caption{A rationale generated by GPT4 with COT shows that even though the model is able to cite the relevant information, it inclines to answer ``No" because it cannot ``definitively conclude".}
    \begin{tabular}{p{.2\textwidth}p{.6\textwidth}}
        \toprule
        Input & \makecell[tl]{\tf{Premise}: PersonX is of age 17 years \\ 8 months. \\ \tf{Conclusion}: PersonX cannot consume \\ Bourbon.\\ Does premise entail conclusion?}\\
        \addlinespace
        Label & Yes \\
        \addlinespace
        Answer & No \\
        \addlinespace
        Rationale & \makecell[tl]{In many countries, the legal drinking \\ age is 18 or 21. If we do not know the \\ specific country PersonX is in, we \\ cannot definitively conclude that they \\cannot consume Bourbon.}\\
        \bottomrule
    \end{tabular}
    \label{tab:rationale}
\end{table}

\section{Domain-wise statistics of \dataset}
\label{app:statistics}

\Tabref{tab:domain_stats} shows the detailed statistics of \dataset.

    \begin{table}[ht]
        \centering
        \small
        \caption{The number of symbolic rules and knowledge statements in different domains in \dataset.}
        \begin{tabular}{cccc}
        \toprule
        & Rule & Head & Long-tail \\
        \midrule
        Temporal & 81 & 15,143 & 15,317 \\
        \makecell[tc]{Outcomes and Effects} & 132 & 14,966 & 15,010 \\
        \makecell[tc]{Natural Properties} & 139 & 16,788 & 16,669 \\
        Locational & 65 & 7,323 & 7,370 \\
        Total & 417 & 54,220 & 54,366 \\
        \bottomrule
        \end{tabular}
        \label{tab:domain_stats}
    \end{table}

\subsection{Domain-wise human evaluation}
As mentioned in \Ssecref{subsec:evaluation}, we uniformly sample 4,000 statements from \dataset~for human evaluation.~\Tabref{tab:human_eval_domain} provides more detailed domain-wise statistics on the data type conformity and factual correctness performance of \dataset~long-tail knowledge generation. While ``Natural Properties'' has the highest overall accuracy and factuality, model performance on positive templates in~\Tabref{tab:probing} is the lowest while model performance on negative templates is the highest in this domain. This suggests that these LLMs might have been most aligned in this domain during pre-training.

    \begin{table}[ht]
        \centering
        \small
        \caption{The factual and data type accuracy of each domain in human evaluation.}
        \begin{tabular}{cccc}
        \toprule
           & Data Type & Factuality & Overall \\
           \midrule
           Temporal & 90.18 & 85.27 & 77.38\\
           \makecell[tc]{Outcomes and Effects} & 94.97 & 80.73 & 75.98\\
           Natural Properties & 96.61 & 96.61 & 93.81\\
           Locational & 98.88 & 70.79 & 70.79\\ 
        \bottomrule
        \end{tabular}
        \label{tab:human_eval_domain}
    \end{table}

\subsection{Rule definitions}
\label{app:rule_definition}
\Tabref{tab:distribution_plots_rule} shows the definitions of the 6 sampled rules.
\begin{table*}[ht]
    \centering
    \small
    \caption{Rule definitions of six sampled rules.}
    \begin{tabular}{p{.1\textwidth}p{.7\textwidth}}
        \toprule
        Rule0 & \makecell[tl]{
            lived\_in(Person P, Geographic Location A) 
            \&  lived\_during(Person P, Historical Time \\ Period D)  \& existed\_during(Geographic Location A,  Historical Time Period D) \& \\ 
            was\_invented\_in(Product or Technology C, Year Y) \&
           is\_more\_than\_a\_century\_earlier\_\\than(Historical Time Period D, Year Y) \\
            $\to$ is\_not\_able\_to\_use(Person P, Product or Technology C) \\
            } \\
        \midrule
        Rule32 & \makecell[tl]{has\_trouble\_lifting(Person X, Name of Appliance B)
        \& is\_heavier\_than(Object A, \\Name of Appliance B) \\
        $\to$ cannot\_lift(Person X, Object A)}\\
        \midrule
        Rule46 & \makecell[tl]{is\_allergic\_to(Person A, Substance X) 
        \& includes(Name of Cosmetics B, Substance X) \\
        $\to$ cannot\_use(Person A, Name of Cosmetics B)} \\
        \midrule
        Rule88 & \makecell[tl]{died\_in(Historical Figure A, Historical Time Period X) 
        \& was\_created\_during(Artifact B, \\ Historical Time Period Y)
        \& is\_earlier\_than( Historical Time Period X, Historical Time \\ Period Y) \\ 
        $\to$ cannot\_create(Historical Figure A,  Artifact B)} \\
        \midrule
        Rule112 & \makecell[tl]{has\_trouble\_containing(Room B, Furniture C)
        \& is\_larger\_than(Furniture A, Furniture C) \\
        $\to$ cannot\_fit\_in(Furniture A, Room B)} \\
        \midrule
        Rule122 & \makecell[tl]{has\_trouble\_containing(Trunk B, Furniture C)
        \& is\_larger\_than(Furniture A, Furniture C) \\
        $\to$ cannot\_fit\_in(Furniture A, Trunk B)} \\
        \bottomrule
    \end{tabular}
    \label{tab:distribution_plots_rule}
\end{table*}

\section{Amazon Mechanic Turk}

\subsection{Recruiting Workers}

We recruit workers from all English-speaking countries (US, UK, New Zealand, Australia, Canada), although AMT workers are mostly US-based. We use a qualification task to recruit AMT workers. In the qualification task, all workers will be presented with three manually selected statements, which are clear and representative. Each statement has five related questions as described in Appendix~\ref{app:filter_job}. Only workers who answer all the questions correctly will be recruited. In the end, we recruited 38 workers to evaluate the quality of generation and 17 workers as human baselines for the entailment classification task. We paid the workers \$0.47 per annotation for evaluating the quality of generations and \$0.11 per annotation for the entailment classification task, to match \$15 per hour based on their working time.

\subsection{Templates}
\label{app:amt_template}
\begin{figure*}
\begin{minipage}{\linewidth}
      \centering
      \begin{minipage}{0.45\linewidth}
          \begin{figure}[H]
            \captionsetup{width=0.9\textwidth}
              \includegraphics[width=\linewidth]{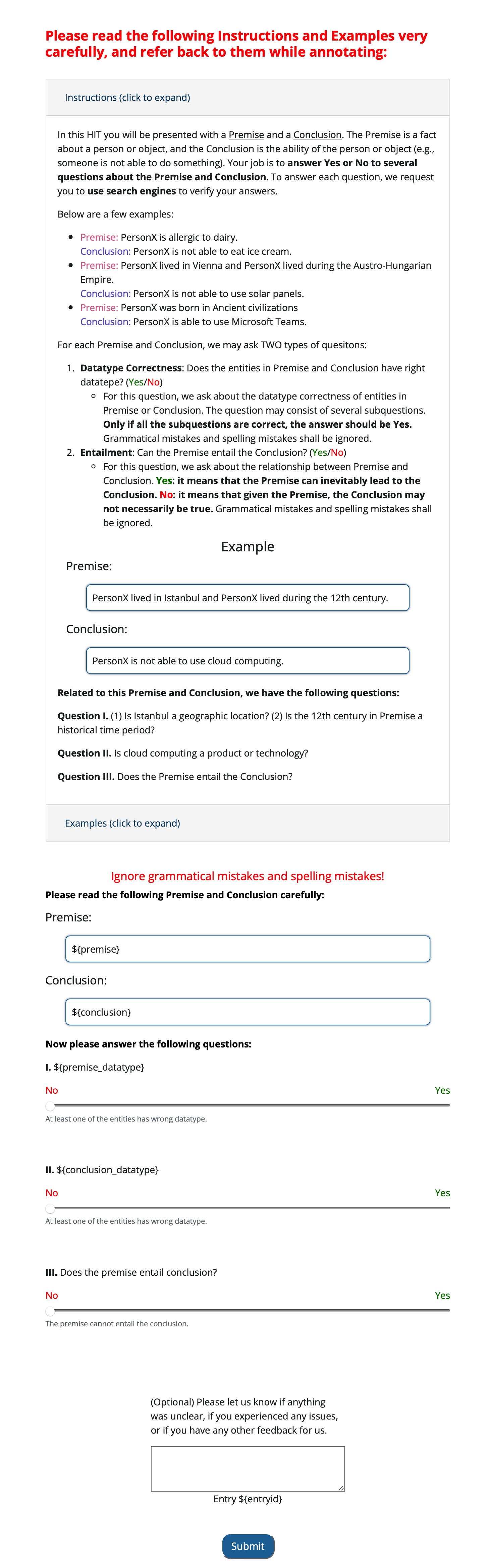}
              \caption{AMT template for the evaluation of generation quality.}
            \label{fig:human_eval_mturk}
          \end{figure}
      \end{minipage}
      \hspace{0.05\linewidth}
      \begin{minipage}{0.45\linewidth}
          \begin{figure}[H]
            \captionsetup{width=0.9\textwidth}
              \includegraphics[width=\linewidth]{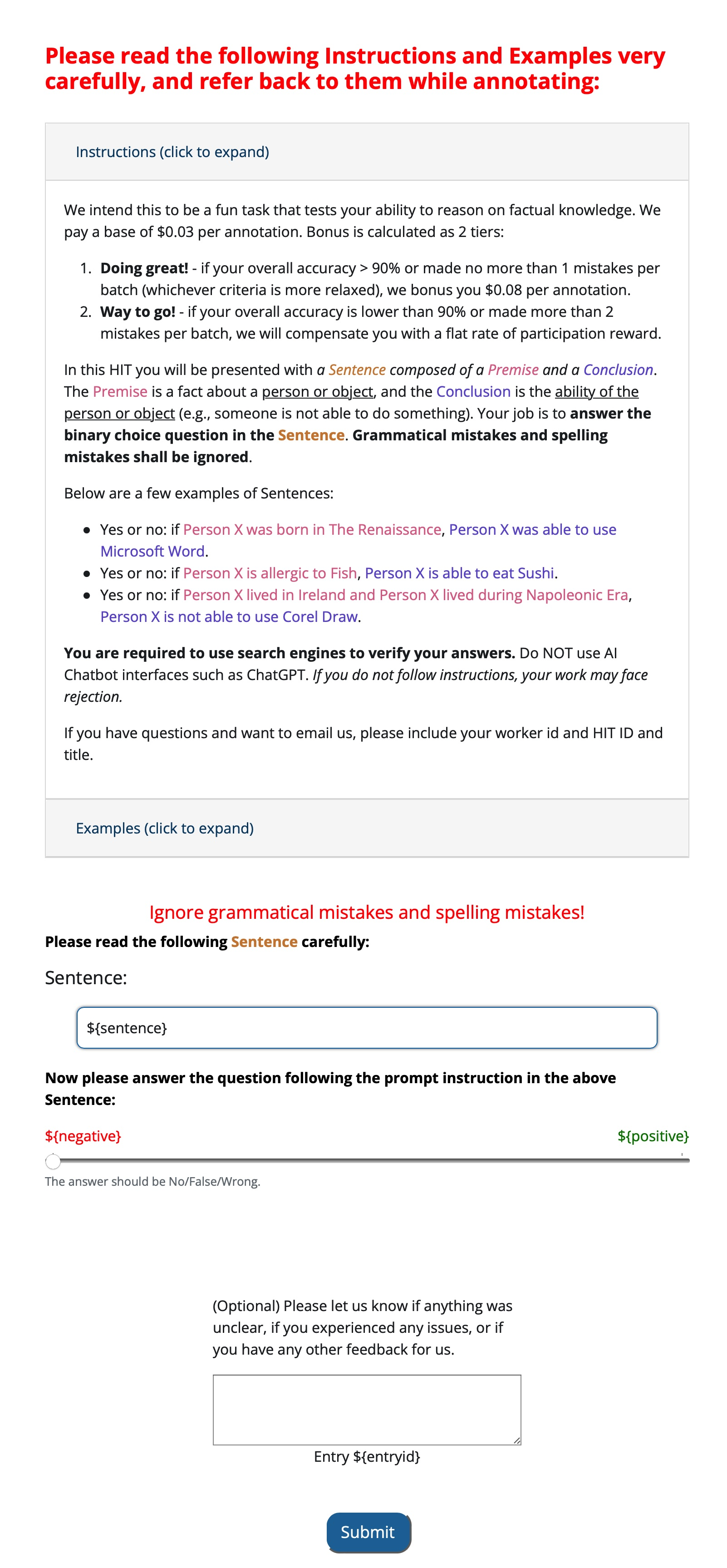}
              \caption{AMT template for the human baseline of the entailment classification task.}
            \label{fig:probing_mturk}
          \end{figure}
      \end{minipage}
\end{minipage}
\end{figure*}

\Figref{fig:human_eval_mturk} and \Figref{fig:probing_mturk} show the template we use for the evaluation of generation quality and the entailment classification task.

\subsection{Agreement statistics}
\label{app:agreement}
\Tabref{tab:agreement} shows the agreement of annotations in the evaluation task. The high agreement of the data type conformity and factual correctness for \dataset~ensures the reliability of our results. The agreement for baselines is lower, which also indicates that the generated statements of baselines are of low quality and confusing for human annotators.
    \begin{table}[ht]
        \centering
        \small
        \caption{Agreement of annotations in the evaluation task.}
        \begin{tabular}{cccc}
        \toprule
          Accuracy & ChatGPT & GPT4 & \framework \\
           \midrule
             Data Type & 79.29 & 83.16 & 87.54 \\
             Factuality & 38.35 & 58.48 & 75.10 \\
             Overall & 65.64 & 74.93 & 83.39 \\
        \bottomrule
        \end{tabular}
        \label{tab:agreement}
    \end{table}

\subsection{Failure Case Examples}
\label{app:failure_cases}

We analyze some failure cases that are labeled as incorrect in the human evaluation. \Tabref{tab:failure_cases} presents some examples.
\begin{table*}[h]
    \centering
    \small
    \caption{Examples that are labeled as incorrect during human evaluation. Note that the reasons are analyzed by the authors instead of annotators.}
    \begin{tabular}{p{.1\textwidth}p{.1\textwidth}p{.6\textwidth}}
        \toprule
        Rule 172
        & Locational & \makecell[tl]{
            \textbf{Rule:} is\_located\_in(Person A, Location X) \& is\_forbidden\_in(Food \\Item B, Location X) $\to$ cannot\_eat(Person A, Food Item B) \\
            \textbf{Premise:} Person X is located in Houston \\
            \textbf{Conclusion:} Person X cannot eat Chocolate \\
            \textbf{Is Houston a location? Annotation:} Yes \\
            \textbf{Is Chocolate a food item? Annotation:} Yes \\
            \textbf{Does the premise entail the conclusion? Annotation:} No\\
            \textbf{Reason:} It is a factual error. Chocolate is not actually forbidden \\in Houston, so People in Houston can eat chocolate. \\
            } \\
        \midrule
        Rule 371
        & Capability and Advice & \makecell[tl]{
            \textbf{Rule:} can\_treat(Drug B, Name of Disease X) \& has(Person A, \\ Name of Disease X) $\to$ should\_take(Person A, Drug B) \\
            \textbf{Premise:} Person X has Hepatitis \\
            \textbf{Conclusion:} Person X should take Sofosbuvir \\
            \textbf{Is Hepatitis a name of disease? Annotation:} Yes \\
            \textbf{Is Sofosbuvir a drug? Annotation:} Yes \\
            \textbf{Does the premise entail the conclusion? Annotation:} No\\
            \textbf{Reason:} It is a factual error. There are different types of hepatitis \\ viruses. Sofosbuvir is a medication used primarily for the treatment \\of hepatitis C. For other types of hepatitis, different medications or \\treatments may be necessary.
            } \\
        \midrule
        Rule 274
        & Temporal & \makecell[tl]{
            \textbf{Rule:} vanished\_in(Plant A, Historical Time Period X) \& \\was\_invented\_in(Weapon B, Historical Time Period Y) \& \\is\_earlier\_than(Historical Time Period X, Historical Time Period Y) \\$\to$ cannot\_be\_used\_to\_conceal(Plant A, Weapon B)\\
            \textbf{Premise:} Plant X vanished in Mongol \\
            \textbf{Conclusion:} Plant X cannot be used to conceal M92 Zolja \\
            \textbf{Is Mongol a historical time period? Annotation:} No \\
            \textbf{Is M92 Zolja a weapon? Annotation:} Yes \\
            \textbf{Does the premise entail the conclusion? Annotation:} Yes\\
            \textbf{Reason:} It is a data type error. The Mongols are an East Asian ethnic \\group native to Mongolia, not a time period. The Mongol Empire may \\ refer to a period of the 13th and 14th centuries, but Mongol cannot.
            } \\
        \midrule
        Rule 204
        & Natural Properties & \makecell[tl]{
            \textbf{Rule: } has\_trouble\_containing(Drawer B, Tool C) \& is\_larger\_than\\(Tool A, Tool C) $\to$ cannot\_be\_placed\_in(Tool A, Drawer B) \\
            \textbf{Premise:} Drawer X has trouble containing Scroll saw \\
            \textbf{Conclusion:} Car cannot be placed in Drawer X \\
            \textbf{Is Scroll saw a Tool? Annotation:} Yes \\
            \textbf{Is Car a Tool? Annotation:} No \\
            \textbf{Does the premise entail the conclusion? Annotation:} Yes\\
            \textbf{Reason:} It is a data type error. Car is a vehicle instead of a tool.
            } \\
        
        \bottomrule
    \end{tabular}
    \label{tab:failure_cases}
\end{table*}

\newpage

\end{document}